%% file: paper.tex
\documentclass{article} 
\usepackage{iclr2018_conference,times}

\usepackage[printonlyused,nohyperlinks]{acronym}
\iclrfinalcopy

\usepackage{amsmath}
\usepackage{amssymb}
\usepackage{graphicx}
\usepackage[colorlinks=true, allcolors=blue]{hyperref}
\usepackage{subcaption}
\usepackage{xspace}
\usepackage{verbatim}

\usepackage{hyperref}
\hypersetup{
    colorlinks=true,
    linkcolor=blue,
    filecolor=magenta,      
    urlcolor=cyan,
}
 
\usepackage[export]{adjustbox}
\usepackage{tikz}
\usetikzlibrary{shapes, arrows, positioning, fit, backgrounds}

\usepackage{siunitx}

\tikzstyle{block} = [rectangle, draw, fill=blue!20, 
    text width=1em, text centered, rounded corners, minimum height=1em, node distance=1.5cm]
\tikzstyle{line} = [draw, -latex']
\tikzstyle{cloud} = [draw, rectangle,fill=red!20, node distance=1.0cm and 1cm,
    minimum height=0.5em]

\input{tex/commands}

\title{Progressive Reinforcement Learning \\ with Distillation \\ for Multi-Skilled Motion Control}
\author{
	Glen Berseth*,
	Cheng Xie*, 
	Paul Cernek,
	Michiel Van de Panne \\
	\texttt{gberseth@cs.ubc.ca,cheng.k.xie@gmail.com,pcernek@cs.ubc.ca,} \\
	\texttt{van@cs.ubc.ca} \\
	University of British Colubia	
}

\newcommand\blfootnote[1]{%
	\begingroup
	\renewcommand\thefootnote{}\footnote{#1}%
	\addtocounter{footnote}{-1}%
	\endgroup
}

\begin{document}
\maketitle

\input{tex/glossary}	

\input{tex/abstract}

\renewcommand{\thefootnote}{\fnsymbol{footnote}}
\blfootnote{* These authors contributed equally to this work.}
\renewcommand{\thefootnote}{\arabic{footnote}}

\input{tex/intro}

\input{tex/related_work}

\input{tex/background}

\input{tex/methods}

\input{tex/results}

\input{tex/discussion}

\bibliographystyle{iclr2018_conference}
\bibliography{paper}

\input{tex/appendix}

\end{document}

%% file: tex/commands.tex
\newcommand{\todo}[1]{\textcolor{red}{To-do: #1}}
\newcommand{\glen}[1]{\textcolor{blue}{Glen: #1}}

\newcommand{\citationneeded}[0]{\textcolor{red}{[citation needed]}}
\definecolor{mygreen}{rgb}{0.058, 0.688, 0.078}
\newcommand{\changes}[1]{#1}

\newcommand{\refSection}[1]{Section: \ref{#1}}
\newcommand{\refAppendix}[1]{Appendix: \ref{#1}}
\newcommand{\refFigure}[1]{Figure~\ref{#1}}
\newcommand{\refTable}[1]{Table~\ref{#1}}

\newcommand{\valueWithUnits}[2]{#1~\scriptsize #2\normalsize}

\newcommand{\normalDistribution}[2]{\mathcal{N}(#1,#2)}
\newcommand{\expectation}{\mathbb{E}}

\newcommand{\grad}[0]{\nabla}
\newcommand{\given}{\:\vert\:}

\DeclareMathOperator*{\argmax}{arg\,max \:}

\newcommand{\deepRL}{\ac{DRL}\xspace}
\newcommand{\HRL}{\ac{HRL}\xspace}

\newcommand{\MDP}{\ac{MDP}\xspace}

\newcommand{\FSM}{\ac{FSM}\xspace}
\newcommand{\TL}{\ac{TL}\xspace}
\newcommand{\RL}{\ac{RL}\xspace}
\newcommand{\MultiTasker}[0]{\ac{MultiTasker}\xspace}
\newcommand{\Parallel}[0]{\ac{Parallel}\xspace}

\newcommand{\PTD}[0]{\ac{PTD}\xspace}
\newcommand{\MSE}[0]{\ac{MSE}\xspace}
\newcommand{\DQN}[0]{\ac{DQN}\xspace}
\newcommand{\SGD}[0]{\ac{SGD}\xspace}
\newcommand{\progRL}[0]{\ac{PLAiD}\xspace}
\newcommand{\PLAID}[0]{\progRL}
\newcommand{\TLOnly}[0]{TL-Only\xspace}

\newcommand{\expert}[0]{\emph{expert}\xspace}
\newcommand{\experts}[0]{\emph{experts}\xspace}
\newcommand{\teacher}[0]{\emph{teacher}\xspace}
\newcommand{\student}[0]{\emph{student}\xspace}
\newcommand{\distillationText}[0]{\emph{distillation}\xspace}
\newcommand{\distilledText}[0]{\emph{distilled}\xspace}
\newcommand{\pdcontrollerText}[0]{\emph{pd-biped}\xspace}

\newcommand{\Flat}{\emph{flat}\xspace}
\newcommand{\incline}{\emph{incline}\xspace}

\newcommand{\gaps}{\emph{gaps}\xspace}

\newcommand{\walls}{\emph{walls}\xspace}
\newcommand{\slopes}{\emph{slopes}\xspace}
\newcommand{\steps}{\emph{steps}\xspace}
\newcommand{\mixed}{\emph{mixed}\xspace}

\newcommand{\stateSpace}{S}
\newcommand{\actionSpace}{A}

\newcommand{\myState}{\sstate}
\newcommand{\sstate}{s}

\newcommand{\action}{a}
\newcommand{\reward}{r}
\newcommand{\ttime}{t}

\newcommand{\discountFactor}{\gamma}

\newcommand{\agent}{agent\xspace}
\newcommand{\character}{character\xspace}

\newcommand{\task}{\omega\xspace}

\newcommand{\taskDomain}{D\xspace}
\newcommand{\taskTarget}{\task_{i+1}\xspace}
\newcommand{\taskSource}{\task_{i}\xspace}

\newcommand{\policySymbol}[0]{\pi}
\newcommand{\policy}[1]{\policySymbol(#1)}

\newcommand{\policyMean}[1]{\mu(#1)}

\newcommand{\valueFunction}[1]{V_{\policySymbol}(#1)}

\newcommand{\AdvantageSymbol}{A}
\newcommand{\advantageFunction}[1]{\AdvantageSymbol_{\policySymbol}(#1)}

\newcommand{\modelParametersPolicy}[0]{\theta_{\mu}}

\newcommand{\modelParametersStocasticPolicy}[0]{\theta_{\policySymbol}}

\newcommand{\modelParametersValueFunction}[0]{\theta_{v}}

%% file: tex/glossary.tex
\acrodef{UI}{user interface}
\acrodef{UBC}{University of British Columbia}
\acrodef{MDP}{Markov Decision Processes}


\acrodef{ANOVA}[ANOVA]{Analysis of Variance\acroextra{, a set of
  statistical techniques to identify sources of variability between groups}}
\acrodef{API}{application programming interface}
\acrodef{CTAN}{\acroextra{The }Common \TeX\ Archive Network}
\acrodef{DOI}{Document Object Identifier\acroextra{ (see
    \url{http://doi.org})}}
\acrodef{GPS}[GPS]{Graduate and Postdoctoral Studies}
\acrodef{PDF}{Portable Document Format}
\acrodef{RCS}[RCS]{Revision control system\acroextra{, a software
    tool for tracking changes to a set of files}}
\acrodef{TLX}[TLX]{Task Load Index\acroextra{, an instrument for gauging
  the subjective mental workload experienced by a human in performing
  a task}}
\acrodef{UML}{Unified Modelling Language\acroextra{, a visual language
    for modelling the structure of software artefacts}}
\acrodef{URL}{Unique Resource Locator\acroextra{, used to describe a
    means for obtaining some resource on the world wide web}}
\acrodef{W3C}[W3C]{\acroextra{the }World Wide Web Consortium\acroextra{,
    the standards body for web technologies}}
\acrodef{XML}{Extensible Markup Language}
\acrodef{MBAE}{Model-Based Action Exploration}
\acrodef{SMBAE}{Stochastic Model-Based Action Exploration}
\acrodef{DRL}{Deep Reinforcement Learning}
\acrodef{HRL}{Hierarchical Reinforcement Learning}
\acrodef{DDPG}{Deep Deterministic Policy Gradient}
\acrodef{CACLA}{Continuous Actor Critic Learning Automaton}
\acrodef{HLC}{High-Level Controller}
\acrodef{LLC}{Low-Level Controller}
\acrodef{ReLU}{Rectified Linear Unit}
\acrodef{FSM}{Finite State Machine}
\acrodef{TL}{Transfer Learning}
\acrodef{MultiTasker}{controller that learns multiple tasks at the same time}
\acrodef{Scratch}{randomly initialized controller}
\acrodef{Parallel}{randomly initialize controllers and train in parallel}
\acrodef{RL}{Reinforcement Learning}
\acrodef{PTD}{Positive Temporal Difference}
\acrodef{MSE}{Mean Squared Error}
\acrodef{DQN}{Deep Q-Network}
\acrodef{SGD}{Stochastic Gradient Decent}
\acrodef{PLAiD}{Progressive Learning and Integration via Distillation}

%

%% file: tex/abstract.tex
\begin{abstract}
Deep reinforcement learning has demonstrated increasing capabilities for continuous control problems,
including agents that can move with skill and agility through their environment. 
An open problem in this setting is that of developing good strategies for integrating or merging policies
for multiple skills, where each individual skill is a specialist in a specific skill and its associated state distribution. 
We extend policy distillation methods to the continuous action setting and leverage this technique to combine \expert policies,
as evaluated in the domain of simulated bipedal locomotion across different classes of terrain.
We also introduce an {\em input injection} method for augmenting an existing policy network to exploit new input features.
Lastly, our method uses transfer learning to assist in the efficient acquisition of new skills.
The combination of these methods allows a policy to be incrementally augmented with new skills.
We compare our progressive learning and integration via distillation (PLAID) method
against three alternative baselines.
\end{abstract}

%% file: tex/intro.tex
\section{Introduction}

As they gain experience, humans develop rich repertoires of motion skills
that are useful in different contexts and environments.
Recent advances in reinforcement learning provide an opportunity to understand how motion 
repertoires can best be learned, recalled, and augmented.
Inspired by studies on the development and recall of movement patterns useful
for different locomotion contexts~\citep{roemmich2015two}, we develop and evaluate an approach
for learning multi-skilled movement repertoires. 
In what follows, we refer to the proposed method as PLAID:  
Progressive Learning and Integration via Distillation. 

For \emph{long lived} applications of complex control tasks 
a learning system may need to acquire and integrate additional skills. 
Accordingly, our problem is defined by the sequential acquisition and integration of new skills. 
Given an existing controller that is capable of one-or-more skills, we wish to: (a) efficiently
learn a new skill or movement pattern in a way that is informed by the existing control policy,
and (b) to reintegrate that into a single controller that is capable of the full motion repertoire.
This process can then be repeated as necessary.  
\changes{We view PLAID as a continual learning method, in that we consider a context where
all tasks are not known in advance and we wish to learn any new task in an efficient manner.
However, it is also proves surprisingly effective as a multitask solution, 
given the three specific benchmarks that we compare against.}
In the process of acquiring a new skill, we also allow for a control policy to be augmented with additional inputs, 
without adversely impacting its performance. This is a process we refer to as {\em input injection}.

Understanding the time course of sensorimotor learning in human motor control 
is an open research problem~\citep{wolpert2016computations} that exists concurrently 
with recent advances in deep reinforcement learning. Issues of generalization,
context-dependent recall, transfer or "savings" in fast learning, forgetting, and scalability
are all in play for both human motor control models and the learning curricula 
proposed in reinforcement learning.   While the development of hierarchical models
for skills offers one particular solution that supports scalability and that avoids  
problems related to forgetting, we eschew this approach in this work and instead
investigate a progressive approach to integration into a control policy defined by a single
deep network.

\textit{Distillation} refers to the problem of combining the policies of one or more \experts in order to create one single controller that can perform the tasks of a set of \experts. 
It can be cast as a supervised regression problem 
where the objective is to learn a model that matches the output distributions of all \expert 	policies~\citep{parisotto2015actor,teh2017distral,rusu2015policy}.
However, given a new task for which an \expert is not given, it is less clear how to learn the new task while successfully integrating this new skill in the pre-existing 
repertoire of the control policy for an \agent.
One well-known technique in machine learning to significantly improve 
sample efficiency across similar tasks is to use \TL~\citep{5288526}, which seeks to 
reuse knowledge learned from solving a previous task to efficiently learn a new task.
However, transferring knowledge from previous tasks to new tasks may not be straightforward; 
there can be \emph{negative transfer} wherein a previously-trained model can take 
longer to learn a new task via fine-tuning than would a randomly-initialized model~\citep{rajendran2017attend}.
Additionally, while learning a new skill, the control policy should not \emph{forget} how to perform old skills.

The core contribution of this paper is a method \PLAID
to repeatedly expand and integrate a motion control repertoire.
The main building blocks consist of policy transfer and multi-task policy \distillationText,
and the method is evaluated in the context of a continuous motor control problem,
that of robust locomotion over distinct classes of terrain.
We evaluate the method against \changes{three} alternative baselines.
We also introduce {\em input injection}, a convenient mechanism for adding inputs
to control policies in support of new skills, while preserving existing capabilities.

%% file: tex/related_work.tex
\section{Related Work}
\label{sec:related-work}

Transfer learning  and distillation are of broad interest in machine learning and RL~\citep{5288526,taylor2009transfer,teh2017distral}. 
Here we outline some of the most relevant work in the area of \deepRL for continuous control environments.

\paragraph{Distillation}
Recent works have explored the problem of combining multiple \expert policies in the reinforcement learning setting. 
A popular approach uses supervised learning to combine each policy by regression over the action distribution. 
This approach yields model compression~\citep{rusu2015policy} as well as a viable method for 
multi-task policy transfer~\citep{parisotto2015actor} on discrete action domains including the Arcade Learning Environment~\citep{bellemare2013arcade}. 
We adopt these techniques and extend them for the case of complex continuous action space tasks and make use of them as building block.

\paragraph{Transfer Learning}
Transfer learning exploits the structure learned from a previous task in learning a new task.
Our focus here is on transfer learning in environments consisting of continuous control tasks. 
The concept of appending additional network structure while keeping the previous structure to reduce \textit{catastrophic forgetting} 
has worked well on Atari games~\citep{rusu2015policy,parisotto2015actor,Rusu2016,chen2015net2net}
Other methods reproduce data from all tasks to reduce the possibility of forgetting how to perform previously learned skills e.g,~\citep{shin2017continual,DBLP:journals/corr/LiH16e}.
Recent work seeks to mitigate this issue using selective learning rates for specific network parameters~\citep{Kirkpatrick28032017}.
A different approach to combining policies is to use a hierarchical structure \citep{Tessler2016}. 
In this setting, previously-learned policies are available as options to execute for a policy trained on a new task. 
However, this approach assumes that the new tasks will be at least a partial composition of previous tasks, 
and there is no reintegration of newly learned tasks.
A recent promising approach has been to apply meta-learning to achieve control policies that can 
quickly adapt their behaviour according to current rewards~\citep{finn2017model}. This work is demonstrated on parameterized task domains.
\changes{The Powerplay method provides a general framework for training an increasingly general problem solver~\citep{DBLP:journals/corr/abs-1112-5309,DBLP:journals/corr/abs-1210-8385}. 
It is based on iteratively: inventing a new task using play or invention; solving this task; 
and, lastly, demonstrating the ability to solve all the previous tasks. 
The last two stages are broadly similar to our PLAID approach, although to the 
best of our knowledge, there are no experiments on motor control tasks of 
comparable complexity to the ones we tackle.  In our work, 
we develop a specific progressive learning-and-distillation methodology for motor skills, 
and provide a detailed evaluation as compared to three other plausible baselines. 
We are specifically interested in understanding issues that arise from the interplay between 
transfer from related tasks and the forgetting that may occur.}

\paragraph{Hierarchical RL} further uses modularity to achieve transfer learning for robotic tasks~\citep{Tessler2016}
This allows for the substitution of network modules for different robot types over a similar tasks~\citep{devin2017learning}. 
Other methods use \HRL as a method for simplifying a complex motor control problem, 
defining a decomposition of the overall task into smaller tasks~\citep{HDeepRL,DBLP:journals/corr/HeessWTLRS16,peng2017deeploco}
While these methods examine knowledge transfer, they do not examine the reintegration of policies 
for related tasks and the associated problems such as \emph{catastrophic forgetting}.
Recent work examines learned motions that can be shaped by prior mocap clips~\citep{merel2017learning}, 
and that these can then be integrated in a hierarchical controller.

%% file: tex/background.tex
\section{Framework}

In this section we outline the details of the \RL framework.
We also give an introduction to the concepts of \TL and \distillationText.

\subsection{Reinforcement Learning}
Leveraging the framework of reinforcement learning, we frame the problem as a \MDP: 
at each time step $ \ttime $, the world (including the \agent) is in a state $ \myState_{\ttime} 
\in \stateSpace $, wherein the agent is able to perform actions $ \action_{\ttime} \in 
\actionSpace $, sampled from a policy $ \policy{\myState_{\ttime}, \action_{\ttime}} = p(\action_{\ttime} | \myState_{\ttime} ) $ 
and resulting in state $ \myState_{\ttime+1} \in \stateSpace $ according to transition 
probabilities $ T(\myState_{\ttime}, \action_{\ttime}, \myState_{\ttime+1}) $. 
Performing action $ \action_{\ttime} $ from state $ 
\myState_{\ttime} $ produces a reward $ \reward_{\ttime} $; the expected cumulative reward 
earned from following some policy $ \policySymbol $ may then be written as:
\begin{equation}
\label{eq:policy-gradient}
 J(\policySymbol) = \expectation_{\reward_{0}, ..., \reward_T} \left[ \sum_{\ttime=0}^{T} \gamma^\ttime \reward_{\ttime} \right]
\end{equation}

\noindent where $ T $ is the time horizon, and $ \discountFactor $ is the 
discount factor, defining the planning horizon length.

The agent's goal is to learn an optimal policy, $ \policySymbol^{*} $, maximizing $ 
J(\policySymbol) $. If the policy has parameters $ \modelParametersStocasticPolicy$, then the goal may 
be reformulated to identify the optimal parameters $ \modelParametersStocasticPolicy^{*} $:

\begin{equation}
\label{eq:policy-optimization}
\modelParametersStocasticPolicy^{*} = \underset{\modelParametersStocasticPolicy}{\argmax} J(\policy{\cdot | \modelParametersStocasticPolicy}) 
\end{equation}

Our policy models a Gaussian distribution with a mean state dependent mean,
$\mu_{\theta_t}(s_t) $.
Thus, our stochastic policy may be formulated as follows:

\begin{equation}
\label{eq:stochastic-policy-def}
\action_{\ttime} \sim \policy{\action_{\ttime} \given \myState_{\ttime},  \modelParametersStocasticPolicy} = 
\normalDistribution{\policyMean{\myState_{\ttime} \given \modelParametersPolicy}}{\Sigma} 
\qquad 
\Sigma = diag\{\sigma_i^2\}
\end{equation}
\noindent where $ \Sigma $ is a diagonal covariance matrix with entries $ \sigma_i^2 $ on the diagonal, similar to \citep{peng2017deeploco}.

To optimize our policy, we use stochastic policy gradient methods, which are well-established family of techniques for reinforcement learning \citep{sutton2000policy}.
The gradient of the expected reward with respect to the policy parameters, $ \grad_{\modelParametersStocasticPolicy} J(\policy{\cdot | \modelParametersStocasticPolicy}) $, is given by:

\begin{equation}
\label{eq:stochastic-policy-optimization}
\grad_{\modelParametersStocasticPolicy} J(\policy{\cdot | \modelParametersStocasticPolicy}) = \int_{\stateSpace} d_{\theta}(\myState) \int_\actionSpace \grad_{\modelParametersStocasticPolicy} \log(\policy{\action, \myState | \modelParametersStocasticPolicy}) \advantageFunction{\myState,\action} \: d\action \: d\myState
\end{equation}

\noindent where $ d_{\theta} = \int_{\stateSpace} \sum_{\ttime=0}^{T} \discountFactor^\ttime p_0(\myState_0) (\myState_0 \rightarrow \myState \given \ttime, \policySymbol_0) \: d\myState_0 $ is the discounted state distribution, $ p_0(\myState) $ represents the initial state distribution, and $ p_0(\myState_0) (\myState_0 \rightarrow \myState \given \ttime, \policySymbol_0) $ models the likelihood of reaching state $ \myState $ by starting at state $ \myState_0 $ and following the policy $ \policy{\action, \myState | \modelParametersStocasticPolicy} $ for $ T $ steps \citep{silver2014deterministic}. 
$ \advantageFunction{\myState,\action} $ represents an advantage function \citep{schulman2016high}. 
In this work, we use the \PTD update proposed by \citep{vanHasselt2012} for $ \advantageFunction{\myState,\action} $:

\begin{equation}
\label{eq:advantage-function}
\advantageFunction{\myState_{\ttime},\action_{\ttime}} = I \left[ \delta_\ttime > 0 \right] = \begin{cases} 1, \quad \delta_\ttime > 0 \\ 0, \quad \text{otherwise} \end{cases}
\end{equation}

\begin{equation}
\label{eq:temporal-difference}
\delta_{\ttime} = \reward_{\ttime} + \discountFactor \valueFunction{\myState_{\ttime+1}} - \valueFunction{\myState_{\ttime}}
\end{equation}

\noindent where $ \valueFunction{\myState} = \expectation \left[ \sum_{\ttime=0}^{T} \discountFactor^{\ttime} \reward_{\ttime} \given \myState_0 = \myState \right] $ is the value function, which gives the expected discounted cumulative reward from following policy $ \policySymbol $ starting in state $ \myState $.
\PTD has the benefit of being insensitive to the advantage function scale.
Furthermore, limiting policy updates in this way to be only in the direction of actions that have a positive advantage has been found to increase the stability of learning \citep{vanHasselt2012}. 
Because the true value function is unknown, an approximation $ \valueFunction{\cdot \given \modelParametersValueFunction} $ with parameters $ \modelParametersValueFunction $ is learned, which is formulated as the regression problem:

\begin{equation}
\label{eq:value-function-regression}
\text{minimize} \underset{\myState_{\ttime}, \reward_{\ttime}, \myState_{\ttime+1}}{\expectation} \left[ \frac{1}{2} \left( y_{\ttime} - \valueFunction{\myState \given \modelParametersValueFunction} \right)^2 \right], \qquad y_{\ttime} = \reward_{\ttime} + \gamma \valueFunction{\myState_{\ttime+1} \given \modelParametersValueFunction}
\end{equation}

\subsection{Policy Distillation}

Given a set of \expert agents that have solved/mastered different tasks we may want to combine the skills of these different \experts into a single multi-skilled agent.
This process is referred to as \distillationText.
\textit{Distillation} does not necessarily produce an optimal mix of the given \experts but instead tries to produce an \expert that best matches the action distributions produced by all \experts.
This method functions independent of the reward functions used to train each \expert.
\emph{Distillation} also scales well with respect to the number of tasks or \experts that are being combined.

\subsection{Transfer Learning}

Given an \expert that has solved/mastered a task we want to reuse that \expert knowledge in order to learn a new task efficiently.
This problem falls in the area of \emph{Transfer Learning}~\citep{5288526}.
Considering the state distribution \expert is skilled at solving, ($\taskDomain_{\taskSource}$ the \emph{source} distribution) it can be advantageous to start learning a \textit{new}, target task $\taskTarget$ with \emph{target} distribution $\taskDomain_{\taskTarget}$ using assistance from the \expert. 
The \agent learning how to solve the \textit{target} task with domain $\taskDomain_{\taskTarget}$ is referred to as the \student.
When the \expert is used to assist the \student in learning the target task it can be referred to as the \teacher.
The success of these methods are dependent on overlap between the $\taskDomain_{\taskSource}$ and $\taskDomain_{\taskTarget}$ state distributions.

%% file: tex/methods.tex
\section{Progressive Learning}

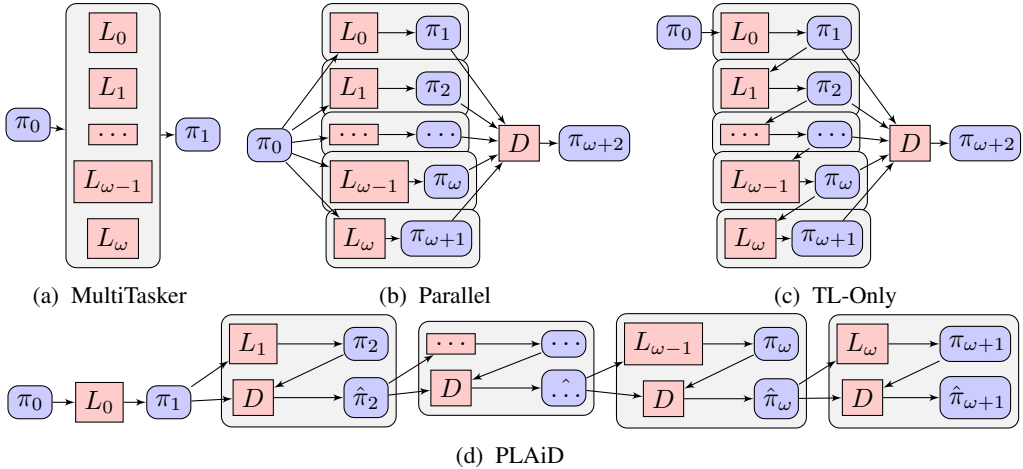
\begin{figure}[h!]
\centering
\subcaptionbox{\label{fig:curriculum-multitasker} \MultiTasker }{
\begin{tikzpicture}[node distance = 0.0cm, auto]
    \node [block] (init) {$ \policySymbol_{0}$};
   
   \node [cloud, above right=-0.5cm and 0.5cm of init] (tl-steps) {$\ldots$};
   \node [cloud, above=0.2cm of tl-steps] (tl-incline) {$L_{1}$};
   \node [cloud, above=0.2cm of tl-incline] (tl-flat) {$L_{0}$};

    \node [cloud, below=0.2cm of tl-steps] (tl-slopes) {$L_{\task - 1}$};
    \node [cloud, below=0.2cm of tl-slopes] (tl-gaps) {$L_{\task}$};
 	\begin{scope}[on background layer]
     \node[draw, rounded corners, fill=gray!10, fit=(tl-flat) (tl-incline) (tl-steps) (tl-slopes) (tl-gaps), inner xsep=1mm] (tl-all){};
     \end{scope}

    \node [block, right=0.5cm of tl-steps] (pi-flat-ramp-steps-slopes) {$ \policySymbol_{1}$};    
    
    \path [line] (init) -> (tl-all);
 
   	
	\path [line] (tl-all) -> (pi-flat-ramp-steps-slopes);

\end{tikzpicture}
}
\centering
\hspace{0.0em}
\subcaptionbox{
\centering
\label{fig:curriculum-distiller-parallel} \Parallel }{
\begin{tikzpicture}[node distance = 0.0cm, auto]
    \node [block] (init) {$ \policySymbol_{0}$};
    
    \node [cloud, above right=1.0cm and 0.5cm of init] (tl-flat) {$L_{0}$};
    \node [block, right=0.5cm of tl-flat] (pi-flat) {$ \policySymbol_{1}$};
    \begin{scope}[on background layer]
    \node[draw, rounded corners, fill=gray!10, fit=(tl-flat) (pi-flat), inner xsep=1mm]{};
    \end{scope}
    
    \node [cloud, below=0.2cm of tl-flat] (tl-incline) {$L_{1}$};
    \node [block, right=0.5cm of tl-incline] (pi-incline) {$ \policySymbol_{2}$};
    \begin{scope}[on background layer]
    \node[draw, rounded corners, fill=gray!10, fit=(tl-incline) (pi-incline), inner xsep=1mm]{};
    \end{scope}
    
    \node [cloud, below=0.2cm of tl-incline] (tl-steps) {$\ldots$};
    \node [block, right=0.5cm of tl-steps] (pi-steps) {$ \ldots$};
    \begin{scope}[on background layer]
	\node[draw, rounded corners, fill=gray!10, fit=(tl-steps) (pi-steps), inner xsep=1mm]{};
	\end{scope}
 
    \node [cloud, below right=0.2cm and -0.65cm of tl-steps] (tl-slopes) {$L_{\task-1}$};
    \node [block, right=0.2cm of tl-slopes] (pi-slopes) {$ \policySymbol_{\task}$};
	\begin{scope}[on background layer]
    \node[draw, rounded corners, fill=gray!10, fit=(tl-slopes) (pi-slopes), inner xsep=1mm]{};
    \end{scope}
    
    \node [cloud, below right=0.2cm and -1.0cm of tl-slopes] (tl-gaps) {$L_{\task}$};
    \node [block, right=0.2cm of tl-gaps, text width=2em] (pi-gaps) {$ \policySymbol_{\task + 1}$};
	\begin{scope}[on background layer]
    \node[draw, rounded corners, fill=gray!10, fit=(tl-gaps) (pi-gaps), inner xsep=1mm]{};
    \end{scope}
    
    \node [cloud, below right=1.0cm and 0.5cm of pi-flat] (u-flat-ramp-steps-slopes) {$D$};
    \node [block, right=0.25cm of u-flat-ramp-steps-slopes, text width=2em] (pi-flat-ramp-steps-slopes) {$\policySymbol_{\task + 2}$};    
    
    \path [line] (init) -> (tl-flat);
    \path [line] (init) -> (tl-incline);
    \path [line] (init) -> (tl-steps);
    \path [line] (init) -> (tl-slopes);
    \path [line] (init) -> (tl-gaps);
    
    \path [line] (tl-flat) -> (pi-flat);
    \path [line] (tl-incline) -> (pi-incline);
	\path [line] (tl-steps) -> (pi-steps);
	\path [line] (tl-slopes) -> (pi-slopes);
    \path [line] (tl-gaps) -> (pi-gaps);
	
	\path [line] (pi-flat) -> (u-flat-ramp-steps-slopes);
	\path [line] (pi-incline) -> (u-flat-ramp-steps-slopes);
	\path [line] (pi-steps) -> (u-flat-ramp-steps-slopes);
	\path [line] (pi-slopes) -> (u-flat-ramp-steps-slopes);
	\path [line] (pi-gaps) -> (u-flat-ramp-steps-slopes);

	\path [line] (u-flat-ramp-steps-slopes) -> (pi-flat-ramp-steps-slopes);	
\end{tikzpicture}
}
\hspace{0.0em}
\subcaptionbox{\label{fig:curriculum-distiller-TL-Only} \TLOnly}{
\begin{tikzpicture}[node distance = 0.0cm, auto]
    \node [block] (init) {$ \policySymbol_{0}$};
    
    \node [cloud, above right=-0.5cm and 0.25cm of init] (tl-flat) {$L_{0}$};
    \node [block, right=0.5cm of tl-flat] (pi-flat) {$ \policySymbol_{1}$};
    \begin{scope}[on background layer]
    \node[draw, rounded corners, fill=gray!10, fit=(tl-flat) (pi-flat), inner xsep=1mm]{};
    \end{scope}
    
    \node [cloud, below=0.2cm of tl-flat] (tl-incline) {$L_{1}$};
    \node [block, right=0.5cm of tl-incline] (pi-incline) {$ \policySymbol_{2}$};
    \begin{scope}[on background layer]
    \node[draw, rounded corners, fill=gray!10, fit=(tl-incline) (pi-incline), inner xsep=1mm]{};
    \end{scope}
    
    \node [cloud, below=0.2cm of tl-incline] (tl-steps) {$\ldots$};
    \node [block, right=0.5cm of tl-steps] (pi-steps) {$ \ldots$};
    \begin{scope}[on background layer]
	\node[draw, rounded corners, fill=gray!10, fit=(tl-steps) (pi-steps), inner xsep=1mm]{};
	\end{scope}
 
    \node [cloud, below right=0.2cm and -0.65cm of tl-steps] (tl-slopes) {$L_{\task-1}$};
    \node [block, right=0.2cm of tl-slopes] (pi-slopes) {$ \policySymbol_{\task}$};
	\begin{scope}[on background layer]
    \node[draw, rounded corners, fill=gray!10, fit=(tl-slopes) (pi-slopes), inner xsep=1mm]{};
    \end{scope}
    
    \node [cloud, below right=0.2cm and -1.0cm of tl-slopes] (tl-gaps) {$L_{\task}$};
    \node [block, right=0.2cm of tl-gaps, text width=2em] (pi-gaps) {$ \policySymbol_{\task + 1}$};
	\begin{scope}[on background layer]
    \node[draw, rounded corners, fill=gray!10, fit=(tl-gaps) (pi-gaps), inner xsep=1mm]{};
    \end{scope}
    
    \node [cloud, below right=1.0cm and 0.5cm of pi-flat] (u-flat-ramp-steps-slopes) {$D$};
    \node [block, right=0.25cm of u-flat-ramp-steps-slopes, text width=2em] (pi-flat-ramp-steps-slopes) {$\policySymbol_{\task + 2}$};    
    
    \path [line] (init) -> (tl-flat);
    
    \path [line] (tl-flat) -> (pi-flat);
    \path [line] (tl-incline) -> (pi-incline);
	\path [line] (tl-steps) -> (pi-steps);
	\path [line] (tl-slopes) -> (pi-slopes);
    \path [line] (tl-gaps) -> (pi-gaps);
    
    \path [line] (pi-flat) -> (tl-incline);
    \path [line] (pi-incline) -> (tl-steps);
   	\path [line] (pi-steps) -> (tl-slopes);
   	\path [line] (pi-slopes) -> (tl-gaps);
	
	\path [line] (pi-flat) -> (u-flat-ramp-steps-slopes);
	\path [line] (pi-incline) -> (u-flat-ramp-steps-slopes);
	\path [line] (pi-steps) -> (u-flat-ramp-steps-slopes);
	\path [line] (pi-slopes) -> (u-flat-ramp-steps-slopes);
	\path [line] (pi-gaps) -> (u-flat-ramp-steps-slopes);

	\path [line] (u-flat-ramp-steps-slopes) -> (pi-flat-ramp-steps-slopes);	
\end{tikzpicture}
}

\subcaptionbox{\label{fig:curriculum-distiller-sequential} \progRL}{
\begin{tikzpicture}[node distance = 0.8cm, auto]
    \node [block] (init) {$ \policySymbol_{0}$};
    \node [cloud, right=0.3cm of init] (tl-flat) {$L_{0}$};
    \node [block, right=0.3cm of tl-flat] (pi-flat) {$ \policySymbol_{1}$};
    
    \node [cloud, above right=0.28cm and 0.5cm of pi-flat] (tl-ramp) {$L_{1}$};
    \node [block, right of=tl-ramp] (pi-ramp) {$ \policySymbol_{2}$};
    \node [cloud, below=0.2cm of tl-ramp] (u-flat-ramp) {$ D $};
    \node [block, right of=u-flat-ramp] (pi-flat-ramp) {$ \hat{\policySymbol}_{2}$};
    \begin{scope}[on background layer]
    \node[draw, rounded corners, fill=gray!10, fit=(tl-ramp) (pi-ramp) (u-flat-ramp) (pi-flat-ramp), inner xsep=1mm]{};
    \end{scope}
        
    \node [cloud, right=0.5cm of pi-ramp] (tl-steps) {$\ldots$};
    \node [block, right of=tl-steps] (pi-steps) {$ \ldots$};
    \node [cloud, below=0.2cm of tl-steps] (u-flat-ramp-steps) {$ D $};
    \node [block, right of=u-flat-ramp-steps] (pi-flat-ramp-steps) {$ \hat{\ldots}$};
    \begin{scope}[on background layer]
    \node[draw, rounded corners, fill=gray!10, fit=(tl-steps) (pi-steps) (u-flat-ramp-steps) (pi-flat-ramp-steps), inner xsep=1mm]{};
    \end{scope}
    
    \node [cloud, right=0.5cm of pi-steps] (tl-slopes) {$L_{\task - 1}$};
    \node [block, right of=tl-slopes] (pi-slopes) {$ \policySymbol_{\task }$};
    \node [cloud, below=0.2cm of tl-slopes] (u-flat-ramp-steps-slopes) {$ D $};
    \node [block, right of=u-flat-ramp-steps-slopes] (pi-flat-ramp-steps-slopes) {$\hat{\policySymbol}_{\task}$};
    \begin{scope}[on background layer]
    \node[draw, rounded corners, fill=gray!10, fit=(tl-slopes) (pi-slopes) (u-flat-ramp-steps-slopes) (pi-flat-ramp-steps-slopes), inner xsep=1mm]{};
    \end{scope}
    
    \node [cloud, right=0.5cm of pi-slopes] (tl-gaps) {$L_{\task }$};
    \node [block, right of=tl-gaps, text width=2em] (pi-gaps) {$ \policySymbol_{\task + 1}$};
    \node [cloud, below=0.2cm of tl-gaps] (u-flat-ramp-steps-slopes-gaps) {$ D $};
    \node [block, right of=u-flat-ramp-steps-slopes-gaps, text width=2em] (pi-flat-ramp-steps-slopes-gaps) {$\hat{\policySymbol}_{\task + 1}$};
    \begin{scope}[on background layer]
    \node[draw, rounded corners, fill=gray!10, fit=(tl-gaps) (pi-gaps) (u-flat-ramp-steps-slopes-gaps) (pi-flat-ramp-steps-slopes-gaps), inner xsep=1mm]{};
    \end{scope}

    \path [line] (init) -> (tl-flat);
    \path [line] (tl-flat) -> (pi-flat);
    
    \path [line] (pi-flat) -> (tl-ramp);
    \path [line] (tl-ramp) -> (pi-ramp);
    \path [line] (pi-flat) -> (u-flat-ramp);
    \path [line] (pi-ramp) -> (u-flat-ramp);
    \path [line] (u-flat-ramp) -> (pi-flat-ramp);
    
    \path [line] (pi-flat-ramp) -> (tl-steps);
    \path [line] (tl-steps) -> (pi-steps);
    \path [line] (pi-flat-ramp) -> (u-flat-ramp-steps);
    \path [line] (pi-steps) -> (u-flat-ramp-steps);
    \path [line] (u-flat-ramp-steps) -> (pi-flat-ramp-steps);
    
    \path [line] (pi-flat-ramp-steps) -> (tl-slopes);
    \path [line] (tl-slopes) -> (pi-slopes);
    \path [line] (pi-flat-ramp-steps) -> (u-flat-ramp-steps-slopes);
    \path [line] (pi-slopes) -> (u-flat-ramp-steps-slopes);
    \path [line] (u-flat-ramp-steps-slopes) -> (pi-flat-ramp-steps-slopes);
    
    \path [line] (pi-flat-ramp-steps-slopes) -> (tl-gaps);
    \path [line] (tl-gaps) -> (pi-gaps);
    \path [line] (pi-flat-ramp-steps-slopes) -> (u-flat-ramp-steps-slopes-gaps);
    \path [line] (pi-gaps) -> (u-flat-ramp-steps-slopes-gaps);
    \path [line] (u-flat-ramp-steps-slopes-gaps) -> (pi-flat-ramp-steps-slopes-gaps);
\end{tikzpicture}
}
\caption{
Different curriculum learning process.
The red box with a $ D $ in it denotes a \distillationText step that combines policies.
Each gray box denotes one iteration of learning a new policy.
The larger red boxes with an $L_{terrain-type}$ denotes a learning step where a new skill is learned.
}
\label{fig:curriculum-flow-chart}
\end{figure}

Although we focus on the problem of being presented with tasks sequentially, there exist other methods for learning a multi-skilled character.
We considered $4$ overall integration methods for learning multiple skills, the first being a \MultiTasker, where a number of skills are learned at the same time.
It has been shown that learning many tasks together can be faster than learning each task separately~\citep{parisotto2015actor}.
The curriculum for using this method is shown in~\refFigure{fig:curriculum-multitasker} were during a single \RL simulation all tasks are learned together.
It is also possible to \changes{\Parallel} and then combine the resulting policies~\refFigure{fig:curriculum-distiller-parallel}.
We found that learning many skills from scratch was challenging, we were only able to get fair results for the \Flat task.
Also, when a \textit{new} task is to be learned with the \changes{\Parallel} model it would occur outside of the original parallel learning, leading to a more sequential method.
A \TLOnly method that uses \TL while learning tasks in a sequence~\refFigure{fig:curriculum-distiller-TL-Only}, possibly ending with a \distillationText step to combine the learned policies to decrease forgetting.
For more details see~\refAppendix{subsection:tl-only-baseline}.
The last version \changes{(\progRL)} learns each task sequentially using \TL from the previous, most skilled policy\changes{, in the end resulting in a policy capable of solving all tasks}~\refFigure{fig:curriculum-distiller-sequential}.
This method works well for both combining learned skills and learning new skills.

\subsection{Progressive Learning and Integration via Distillation}

In this section, we detail our proposed learning framework for continual policy transfer and \distillationText (\progRL). 
In the acquisition (\TL) step, we are interested in learning a new task $\task_{i+1}$. 
Here transfer can be beneficial if the task structure is somewhat similar to previous tasks $\task_{i}$. 
We adopt the \TL strategy of using an existing policy network and fine-tuning it to a new task. 
Since we are not concerned with retaining previous skills in this step, we can update this policy without concern for forgetting.
As the \agent learns it will develop more skills and the addition of every new skill can increase the probability of transferring knowledge to assist the learning of the next skill.

In the integration (\distillationText) step, we are interested in combining all past skills $(\policySymbol_{0}, \ldots, \policySymbol_{i})$ with the newly acquired skill $\policySymbol_{i+1}$. 
Traditional approaches have used policy regression where data is generated by collecting trajectories of the \expert policy on a task. 
Training the \student on these trajectories does not always result in robust behaviour.
This poor behaviour is caused by the \student experiencing a different distribution of trajectories than the \expert during evaluation.
To compensate for this distribution difference, portions of the trajectories should be generated by the \student.
This allows the \expert to suggest behaviour that will pull the state distribution of the \student closer to the \expert's.
This is a common problem in learning a model to reproduce a given distribution of trajectories~\citep{DBLP:journals/corr/abs-1011-0686,NIPS2015_5956,DBLP:journals/corr/MartinezBR17,NIPS2016_6099}.
We use a method similar to the DAGGER algorithm~\citep{DBLP:journals/corr/abs-1011-0686} 
which is useful for distilling policies~\citep{parisotto2015actor}. 
\changes{See~\refAppendix{subsection:distillation-details} for more details.}
As our \RL algorithm is an actor-critic method, we also perform regression on the critic by fitting both in the same step.

\subsection{High Level Experiment Design}

The results presented in this work cover a range of tasks that share a similar action space and state space. Our focus is to demonstrate continual learning between related tasks. 
In addition, the conceptual framework allows for extensions that would permit differing state spaces, described later in~\refSection{sec:feature-injection}.

%% file: tex/results.tex
\section{Results}

In this experiment, our set of tasks consists of $5$ different terrains that a 2D humanoid walker (\pdcontrollerText) learns to traverse. 
The humanoid walker is trained to navigate multiple types of terrain including \Flat in (\refFigure{fig:distilation-environemnts-flat}), \incline (\refFigure{fig:distilation-environemnts-incline}), \steps (\refFigure{fig:distilation-environemnts-steps}), \slopes (\refFigure{fig:distilation-environemnts-slopes}), \gaps (\refFigure{fig:distilation-environemnts-gaps}) and a combination of all terrains \mixed (\refFigure{fig:distilation-environemnts-mixed}) on which agents are trained.
The goal in these tasks is to maintain a consistent forward velocity traversing various terrains, while also matching a motion capture clip of a natural human walking gait on flat ground, similar to~\citep{PengP16}. The \pdcontrollerText receives as input both a \character and  (eventually) a terrain state representation, consisting of
the terrains heights of $50$ equally-spaced points in front of the character. 
The action space is 11-dimensional, corresponding to the joints.  Reasonable torque limits are applied, which
helps produce more natural motions and makes the control problem more difficult.
A detailed description of the experimental setup is included in~\refSection{sec:training-details-on-training}.
The tasks are presented to the \agent sequentially and the goal is to progressively learn to traverse all terrain types.

We evaluate our approach against \changes{three} baselines. 
First, we compare the above learning curriculum from learning new tasks in \progRL with learning new tasks in \Parallel. 
This will demonstrate that knowledge from previous tasks can be effectively transferred after \distillationText steps. 
Second, we compare to the \MultiTasker to demonstrate that iterated \distillationText is effective for the retention of learned skills. 
The \MultiTasker is also used as a baseline for comparing learning speed.
Last, a method that performs \TL between tasks and concludes with a \distillationText step is evaluated to illustrate the result of different \TL and \distillationText schedules.
The results of the \progRL controller are displayed in the accompanying~\href{https://youtu.be/_DjHbHCXGk0}{Video}
\footnote{$https://youtu.be/_DjHbHCXGk0$}

\subsection{Transfer Learning}

\begin{figure}[ht!]
\centering
\subcaptionbox{\label{fig:tl-evalution-incline} \incline}{ \includegraphics[trim={0.0cm 0.0cm 0.0cm 0.0cm},clip,width=0.48\columnwidth]{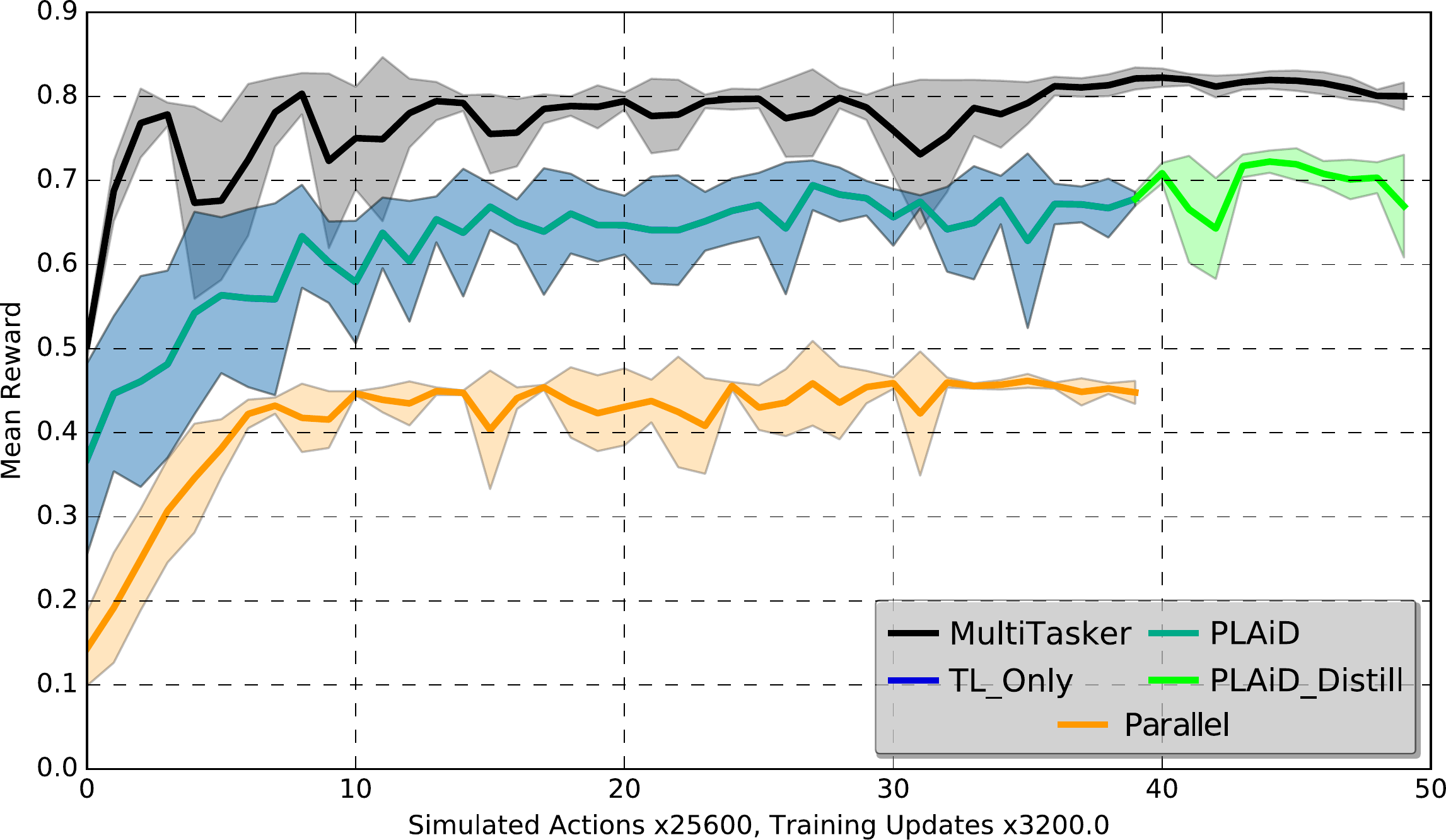}}
\subcaptionbox{\label{fig:tl-evalution-steps} \steps}{ \includegraphics[trim={0.0cm 0.0cm 0.0cm 0.0cm},clip,width=0.48\columnwidth]{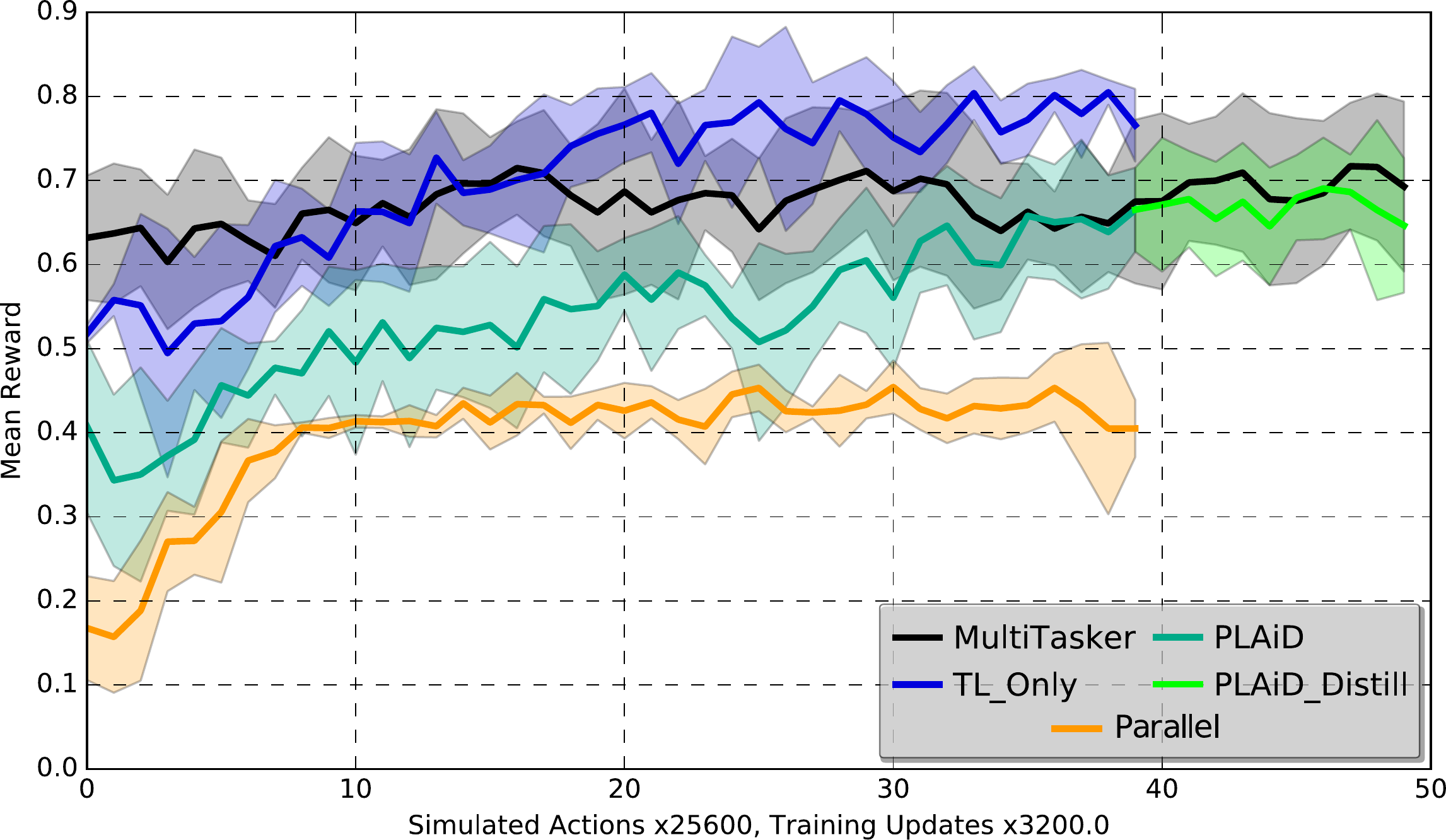}} \\
\subcaptionbox{\label{fig:tl-evalution-slopes} \slopes}{ \includegraphics[trim={0.0cm 0.0cm 0.0cm 0.0cm},clip,width=0.48\columnwidth]{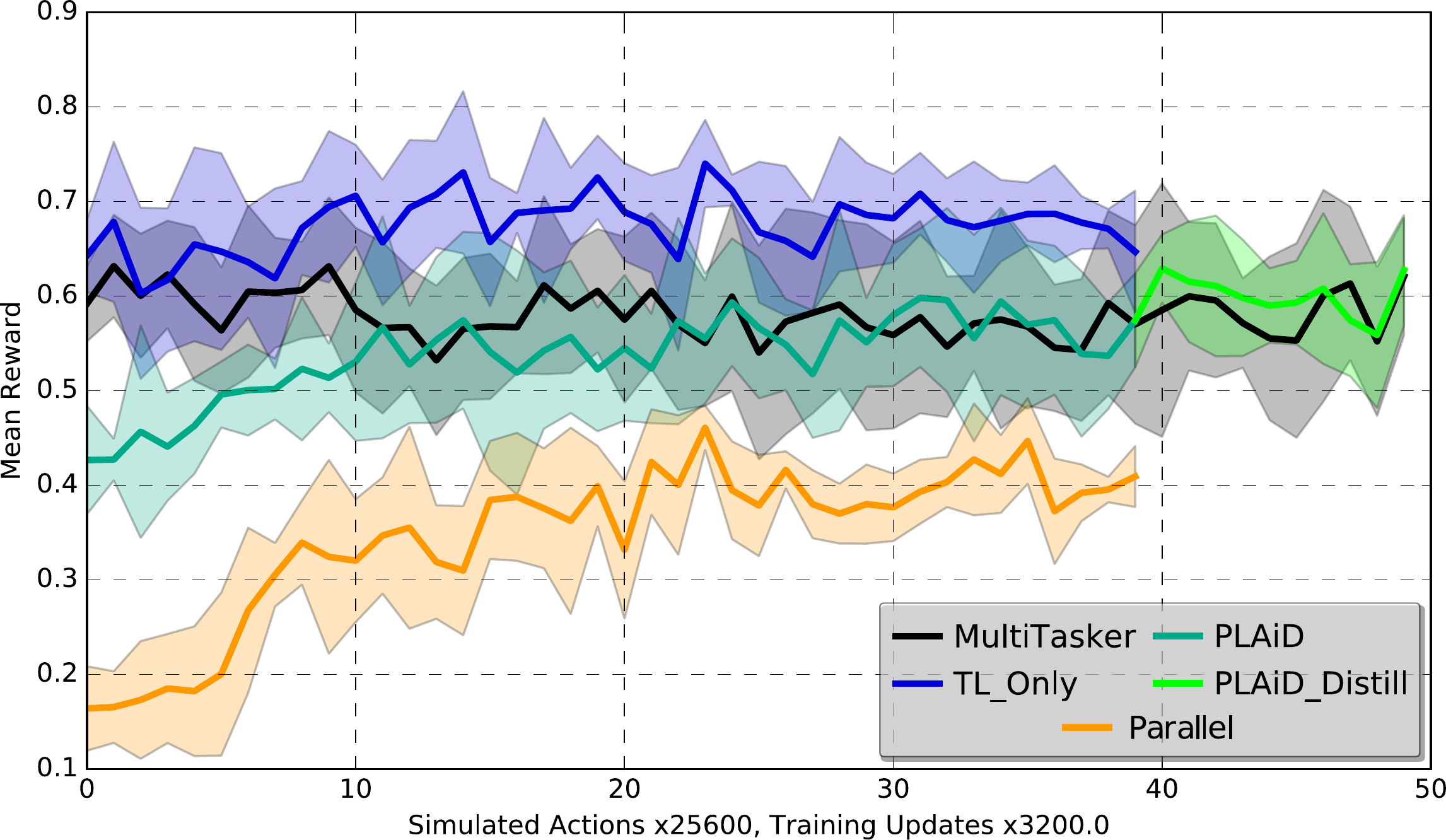}}
\subcaptionbox{\label{fig:tl-evalution-gaps} \gaps}{ \includegraphics[trim={0.0cm 0.0cm 0.0cm 0.0cm},clip,width=0.48\columnwidth]{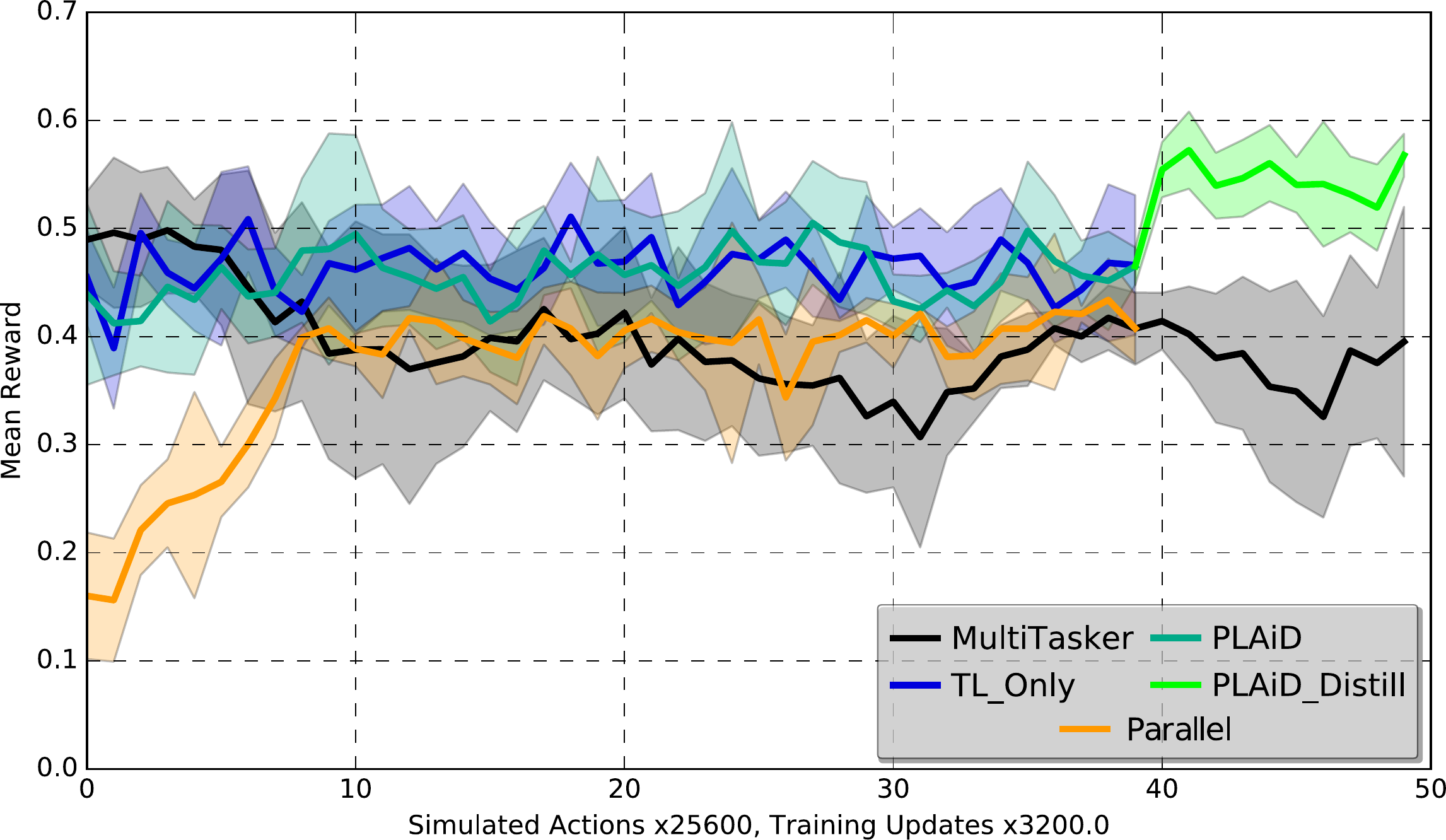}}
\caption{
Learning comparison over each of the environments.
These plots show the mean and std over $5$ simulations, each initialized with different random seeds.
The learning for \progRL is split into two steps, with \TL (in green) going first followed by the \distillationText part (in yellow).
}
\label{fig:distilation-evalution}
\end{figure}

First, the \pdcontrollerText is trained to produce a walking motion on flat ground (\Flat). 
In~\refFigure{fig:tl-evalution-incline} \progRL is compared to the \changes{three} baselines for training on \incline.
The \TLOnly method learns fast as it is given significant information about how to perform similar skills.
The \Parallel method is given no prior information leading to a less skilled policy.
The first \MultiTasker for the \incline task is initialized from a terrain injected controller that was trained to walk on \Flat ground. 
Any subsequent \MultiTasker is initialized from the final \MultiTasker model of the preceding task.
This controller has to learn multiple tasks together, which can complicate the learning process, as simulation for each task is split across the training and the overall \RL task can be challenging.
This is in contrast to using \progRL, that is also initialized with the same policy trained on \Flat,  that will integrate skills together after each new skill is learned.

In~\refFigure{fig:tl-evalution-steps} the \MultiTasker is learning the new task (\steps) with similar speed to \progRL.
However, after adding more tasks the \MultiTasker is beginning to struggle in~\refFigure{fig:tl-evalution-slopes} and starts to \textit{forget} in~\refFigure{fig:tl-evalution-gaps}, with the number of tasks it must learn at the same time.
While \progRL learns the new tasks faster and is able to integrate the new skill required to solve the task robustly.
\TLOnly is also able to learn the new tasks very efficiently.

\subsection{Input Feature Injection}
\label{sec:feature-injection}

An appealing property of using \distillationText in \progRL is that the combined policy model need not resemble that of the individual \expert controllers. For example, two different \experts lacking state features and trained without a local map of the terrain can be combined into a single policy that has new state features for the terrain. 
These new terrain features can assist the \agent in the task domain in which it operates.

We introduce the idea of \emph{input injection} for this purpose. 
We augment a policy with additional input features while allowing it to retain its original functional behaviour similar to \citep{chen2015net2net}.
This is achieved by adding additional inputs to the neural network and initializing the connecting layer weights and biases to $0$. 
By only setting the weights and biases in the layer connecting the new features to the original network to $0$, the gradient can still propagate to any lower layers which are initialized random without changing the functional behaviour.
This is performed when distilling the \Flat and \incline \experts.
\changes{Further details can be found in~\refAppendix{subsection:input-feature-injection}.}

\subsection{Distilling Multiple Policies}

\begin{figure}[htb!]
\centering

\subcaptionbox{\label{fig:distilation-multitask-3} \MultiTasker on $3$ tasks}{ \includegraphics[trim={0.0cm 0.0cm 0.0cm 1.0cm},clip,width=0.32\columnwidth]{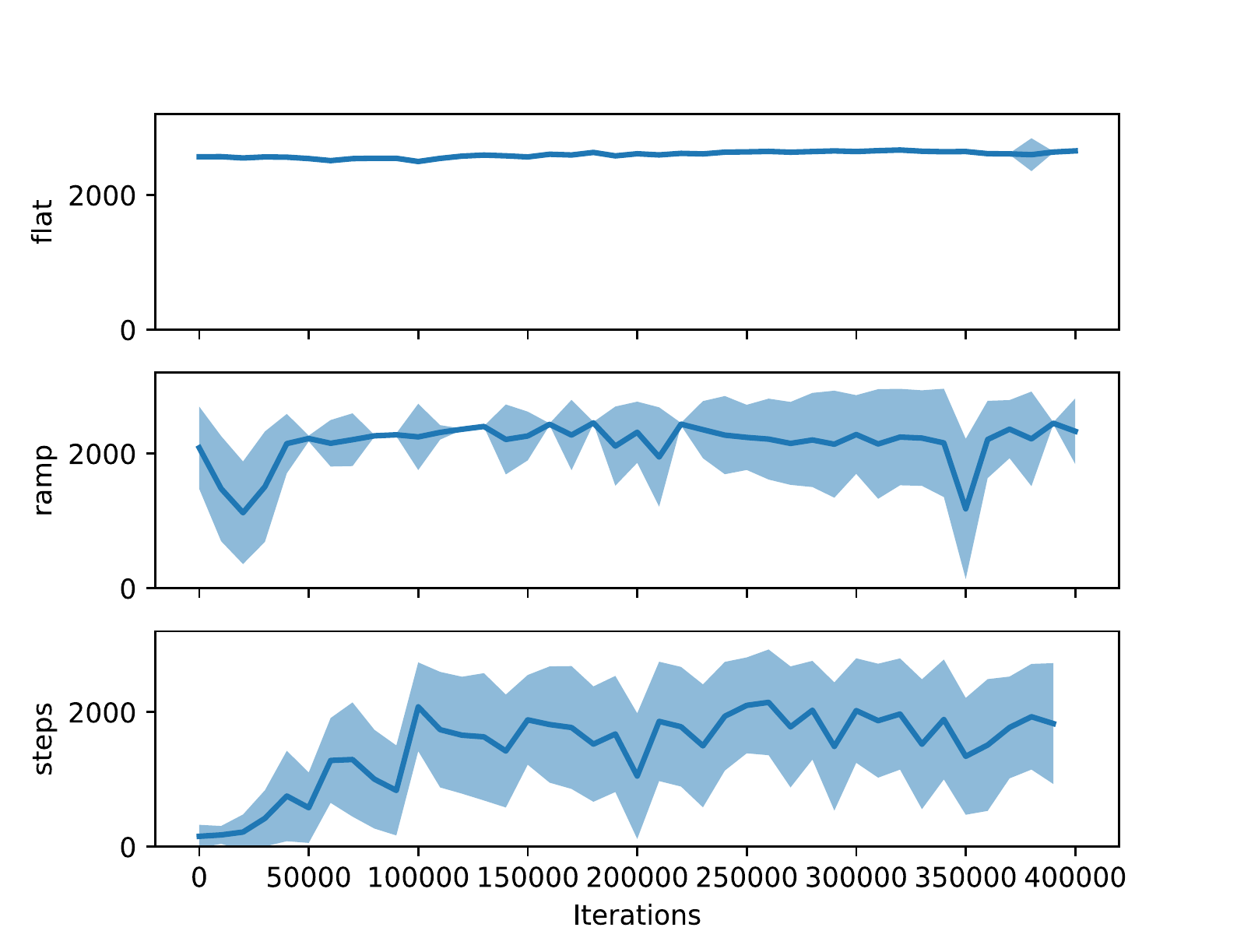}}
\subcaptionbox{\label{fig:distilation-multitask-4} \MultiTasker on $4$ tasks}{ \includegraphics[trim={0.0cm 0.0cm 0.0cm 1.0cm},clip,width=0.32\columnwidth,valign=t]{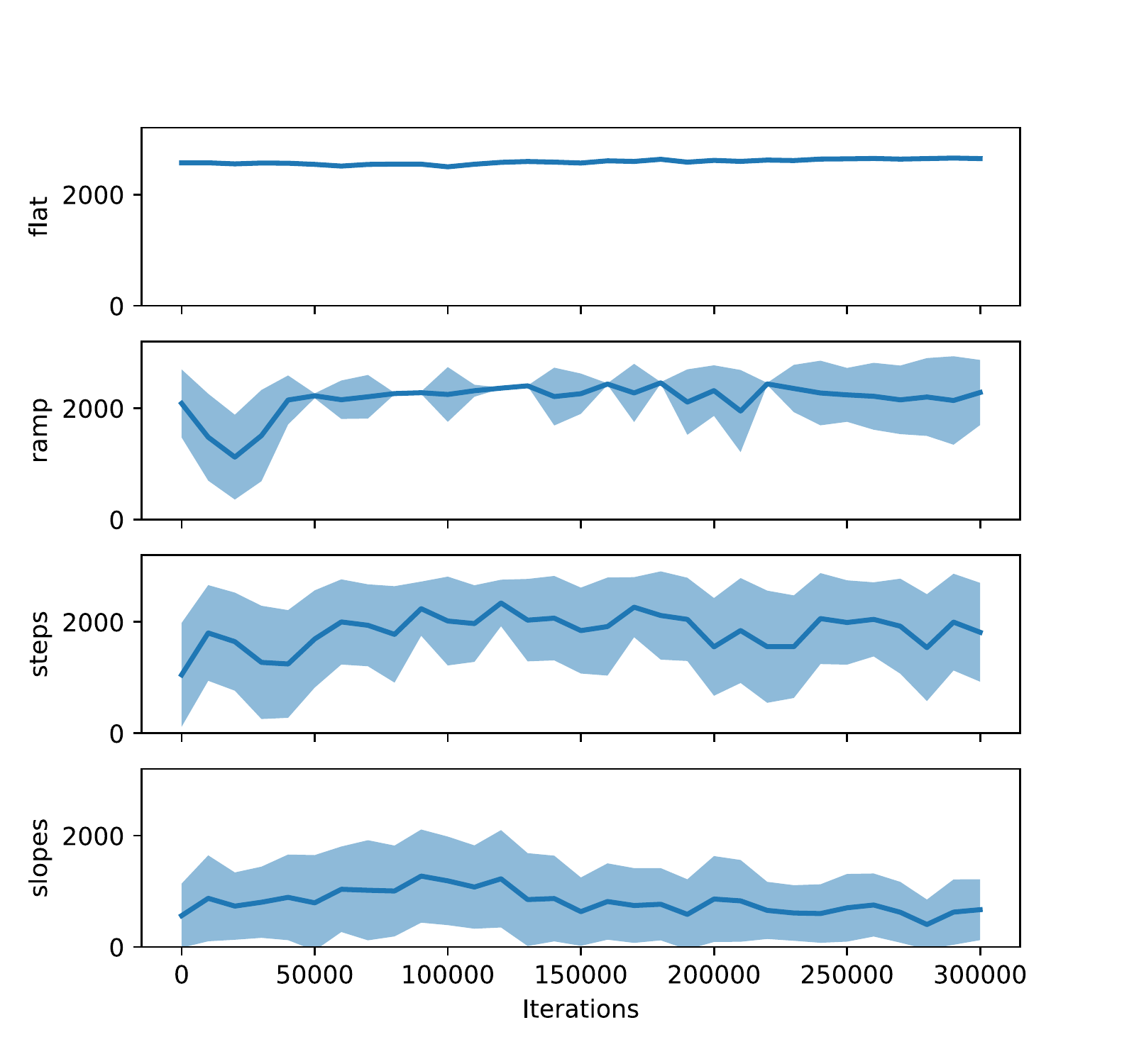}}
\subcaptionbox{\label{fig:distilation-multitask-5} \MultiTasker on $5$ tasks}{ \includegraphics[trim={0.0cm 0.0cm 0.0cm 1.0cm},clip,width=0.32\columnwidth,valign=t]{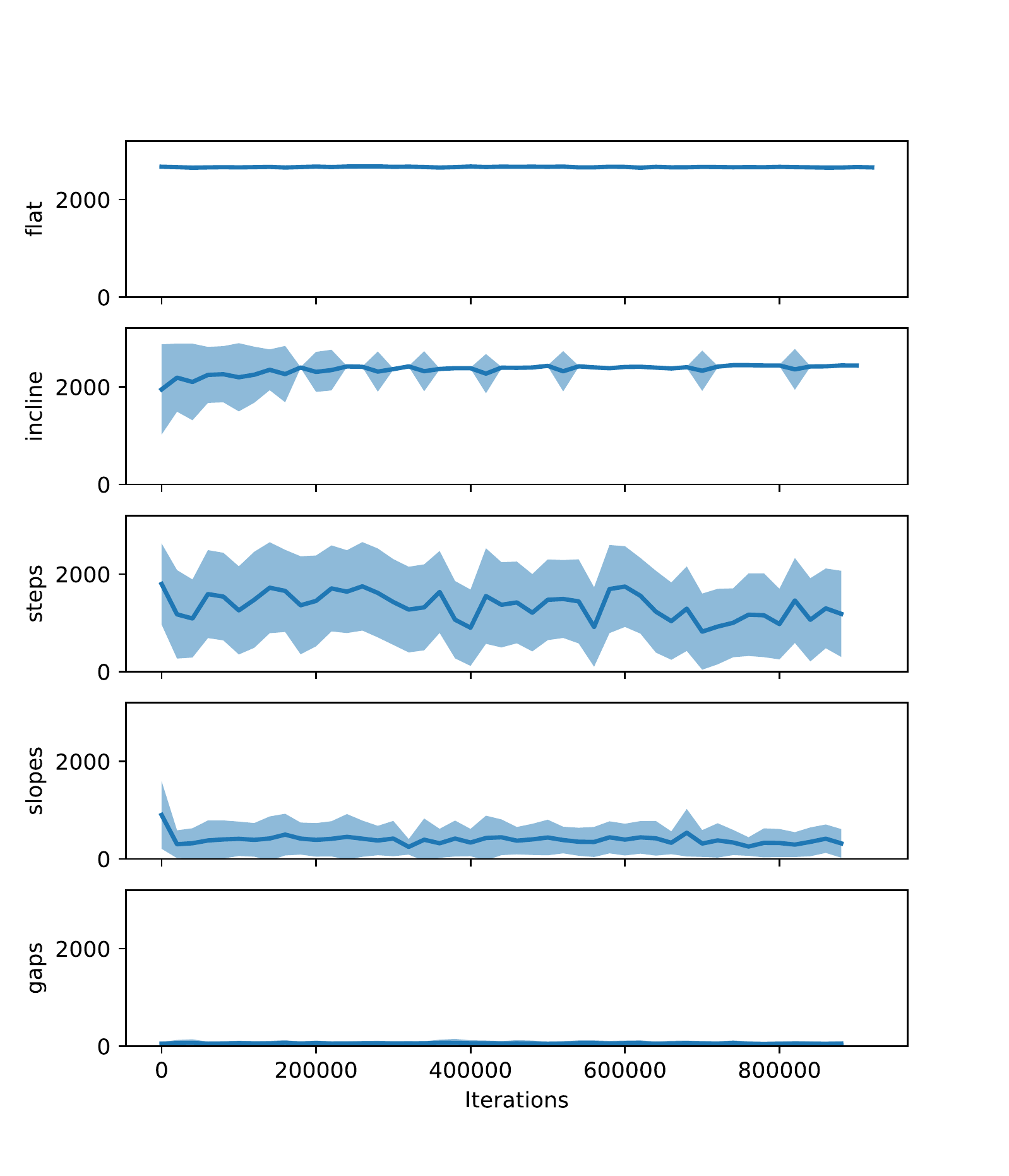}} \\

\subcaptionbox{\label{fig:distilation-distiller-3} \progRL on $3$ tasks}{ \includegraphics[trim={0.0cm 0.0cm 0.0cm 1.0cm},clip,width=0.32\columnwidth,valign=t]{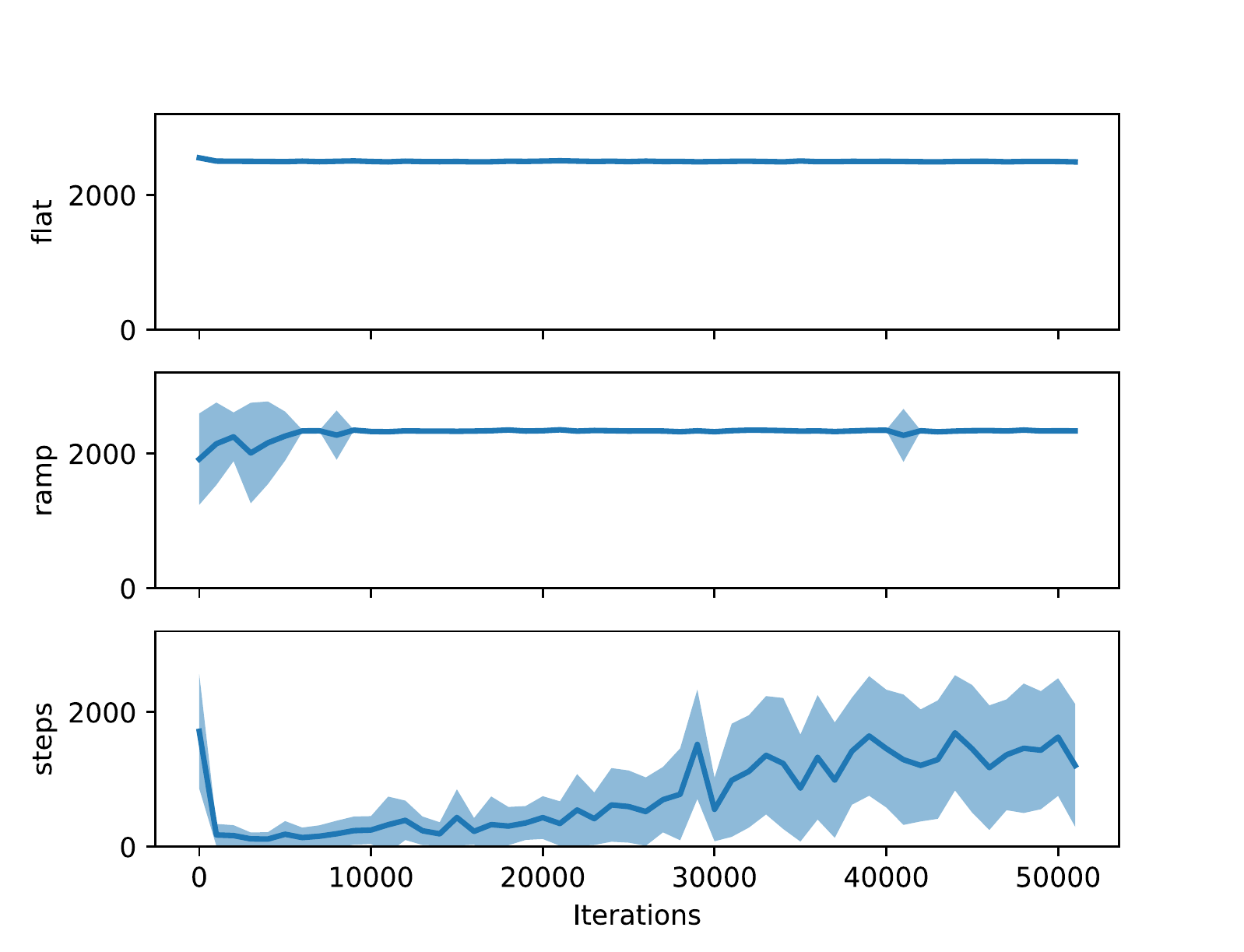}}
\subcaptionbox{\label{fig:distilation-distiller-4} \progRL on $4$ tasks}{ \includegraphics[trim={0.0cm 0.0cm 0.0cm 1.0cm},clip,width=0.32\columnwidth,valign=t]{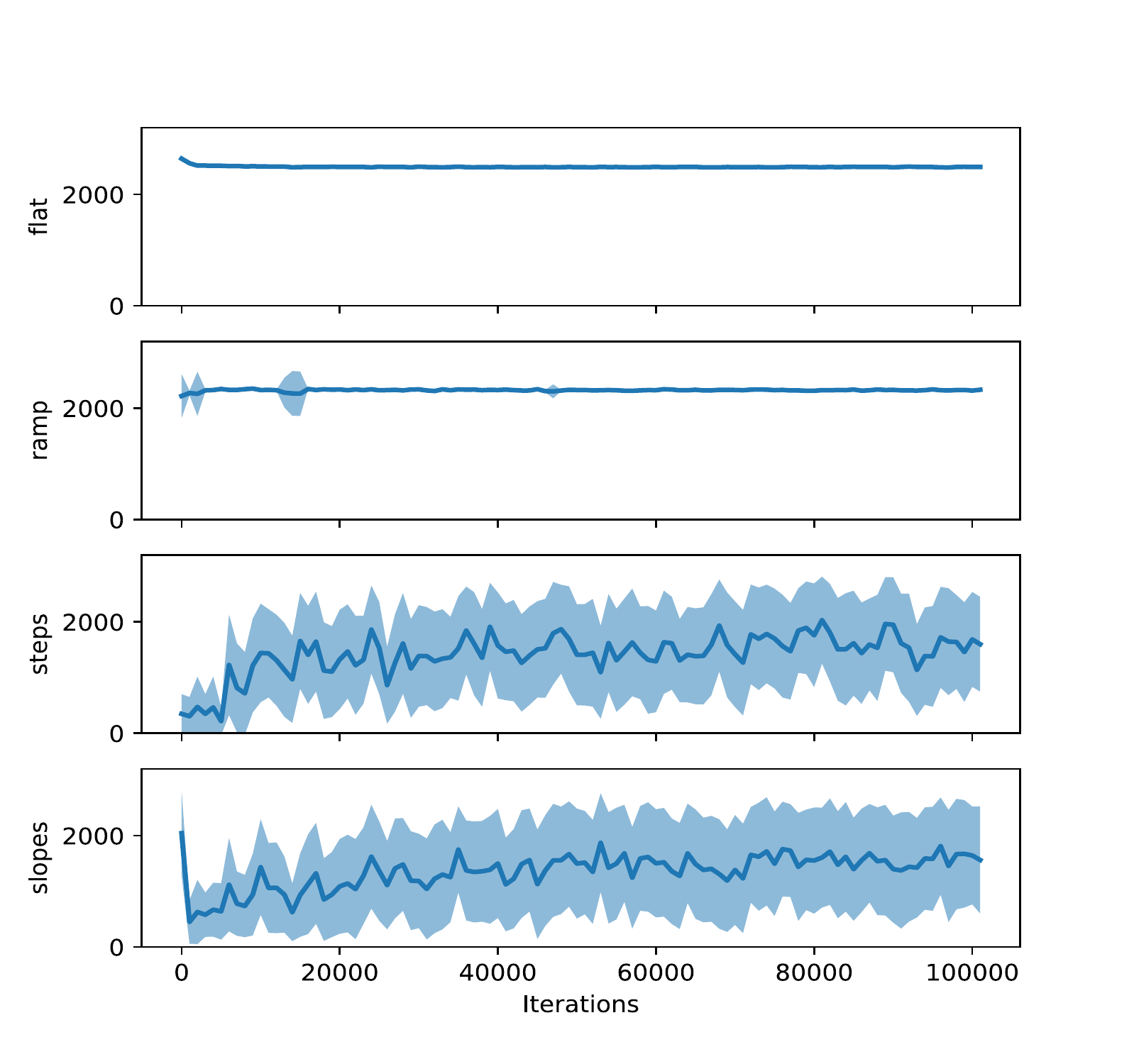}}
\subcaptionbox{\label{fig:distilation-distiller-5} \progRL on $5$ tasks}{ \includegraphics[trim={0.0cm 0.0cm 0.0cm 1.0cm},clip,width=0.32\columnwidth,valign=t]{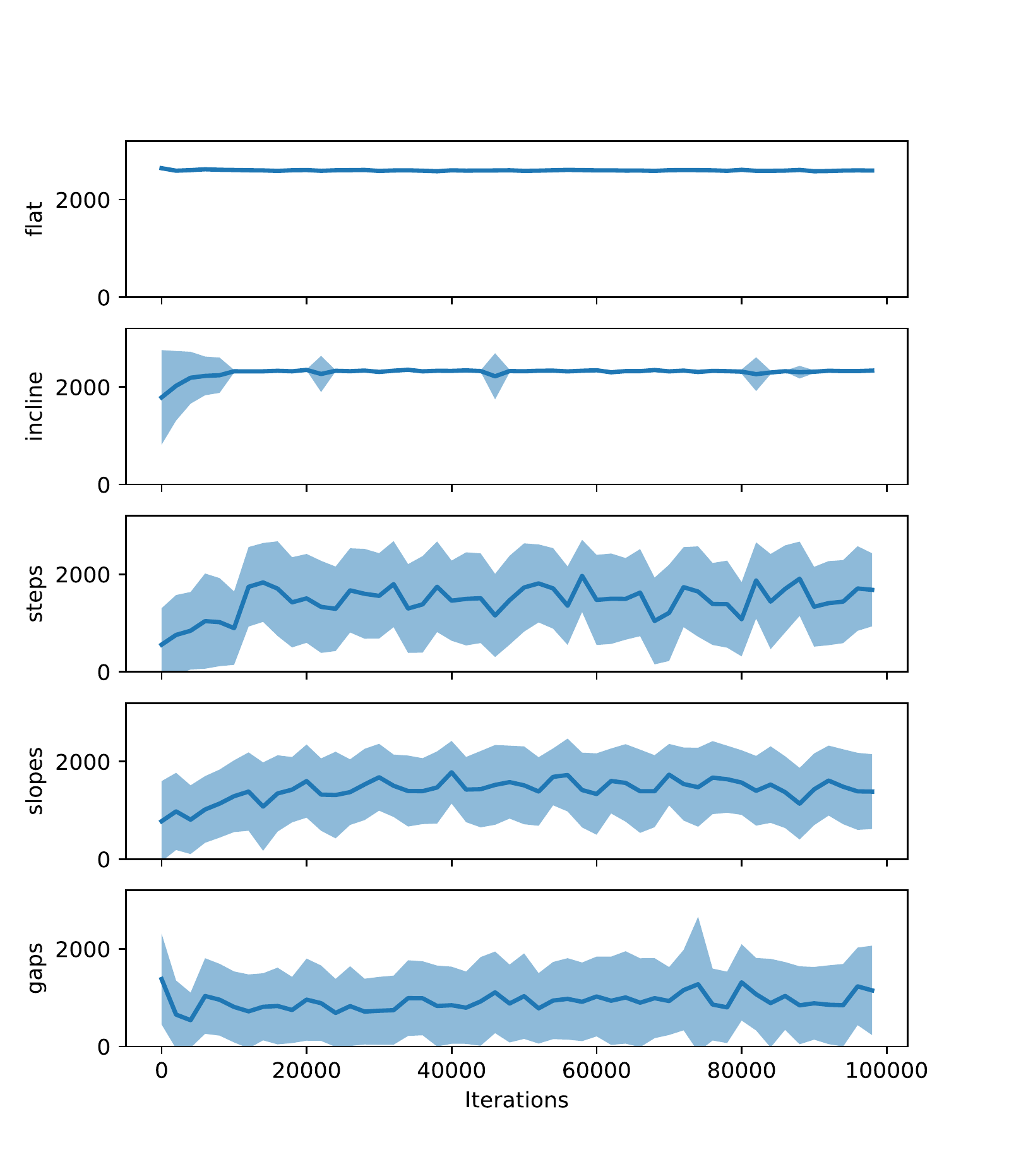}}
\caption{
These figures show the average reward a particular policy achieves over a number of tasks.
 }
\label{fig:distilation-evalution-multitasker}
\end{figure}

Training over multiple tasks at the same time m\changes{a}y help the \agent learn skills quicker, but this may not scale with respect to the number of tasks.
When training the \MultiTasker over two or even three tasks (\refFigure{fig:distilation-multitask-3}) the method displays good results, however when learning a fourth or more tasks the method struggles, as shown in~\refFigure{fig:distilation-multitask-4} and \ref{fig:distilation-multitask-4}.
Part of the reason for this struggle is when new tasks are added the \MultiTasker has to make trade-offs between more tasks to maximizes.
As more tasks are added, this trade-off becomes increasingly complex resulting in the \MultiTasker favouring easier tasks.
Using \progRL to combine the skills of many policies appears to scale better with respect to the number of skills being integrated.
This is likely because distillation is a semi-supervised method which is more stable than the un-supervised \RL solution. 
This can be seen in~\refFigure{fig:distilation-distiller-3}, \ref{fig:distilation-distiller-4} and especially in~\ref{fig:distilation-distiller-5} where \progRL combines the skills faster and can find higher value policies in practice.
\progRL also presents zero-shot training on tasks which it has never \changes{been} trained \changes{on}. In~\refFigure{fig:mixed-environemnts} this generalization is shown as the \agent navigate\changes{s} across the \mixed environment.

\changes{
This is also reflected in~\refTable{fig:TL-only-comparison}, that shows the final average reward when comparing methods before and after distillation.
The \TLOnly is able to achieve high performance but much is lost when learning new tasks. 
A final distillation step helps mitigate this issue but does not work as well as \PLAID.
It is possible performing a large final \distillationText step can lead to over-fitting. 
}

\begin{table}
\centering
\begin{tabular}{|p{3.5cm}||p{1.0cm}|p{1.25cm}|p{1.25cm}|p{1.25cm}|p{1.25cm}|p{1.25cm}|}
\hline
	Tasks & \Flat & \incline & \steps & \slopes & \gaps & \textbf{average} \\ \hline \hline
	\PLAID & \num{0.05369152168} & \num{0.155463332} & \num{0.001460682862} & \textbf{\num{0.04312415806}} &  \num{-0.08282105007} & \num{0.06343492365} \\ \hline
	\TLOnly & \num{-0.064901897} & \num{-0.04426055286} & \num{-0.235345643} & \num{-0.2415014241} & \num{0.0} & \num{-0.1465023792} \\ \hline
    \TLOnly (with Distill) & \num{0.06827925578} & \num{0.03907962212} & \num{-0.02969553903} & \num{-0.06233746537} & \num{-0.1333784127} & \num{-0.02361050785} \\ \hline
    \MultiTasker & \num{-0.001288299624} & \num{-0.05257717165} & \num{-0.02951921005} & \num{0.1190142745} & \num{0.0} & \num{0.008907398302} \\ \hline
\end{tabular}
\caption{
\changes{
These values are relative percentage changes in the average reward, where a value of $0$ is no forgetting and a value of $-1$ corresponds to completely forgetting how to perform the task.
A value $> 0$ corresponds to the agent learning how to better perform a task after training on other tasks.
Here, the final policy after training on \gaps 
compared to the original polices produced at the end of training for the task noted in the column
heading.
The \TLOnly baseline forgets more than \PLAID.
The \MultiTasker forgets less than \PLAID but has a lower average reward over the tasks.}
}
\label{fig:TL-only-comparison}
\end{table}

There are some indications that \distillationText is hindering training during the initial few iterations.
We are initializing the network used in \distillationText with the most recently learning policy after \TL. 
The large change in the initial state distribution from the previous seen distribution during \TL could be causing larger gradients to appear, disrupting some of the structure learned during the \TL step, shown in~\refFigure{fig:distilation-distiller-3} and \ref{fig:distilation-distiller-4}.
There also might not exist a smooth transition in policy space between the newly learned policy and the previous policy distribution.

%% file: tex/discussion.tex
\section{Discussion}

\paragraph{\MultiTasker vs \progRL:}
The \MultiTasker may be able to produce a policy that has higher overall average reward, 
but in practise constraints can keep the method from combining skills gracefully. 
If the reward functions are different between tasks, 
the \MultiTasker can favour a task with higher rewards, as these tasks may receive higher advantage. 
It is also a non-trivial task to normalize the reward functions for each task in order to combine them.
The \MultiTasker may also favour tasks that are easier than other tasks in general.
We have shown that the \PLAID scales better with respect to the number of tasks than the \MultiTasker. 
We expect \progRL would further outperform the \MultiTasker if the tasks were more difficult and the reward functions dissimilar.

In our evaluation we compare the number of iterations \progRL uses to the number the \MultiTasker uses on only the \textit{new} task, 
which is not necessarily fair. The \MultiTasker gains its benefits from training on the other tasks together. 
If the idea is to reduce the number of simulation samples that are needed to learn \textit{new} tasks then the \MultiTasker would fall far behind. 
\textit{Distillation} is also very efficient with respect to the number of simulation steps needed. 
Data could be collected from the simulator in groups and learned from in many batches before more data 
is needed as is common for behavioural cloning. 
We expect another reason \distillationText benefits learning multiple tasks is that
the integration process assists in pulling policies out of the local minima \RL is prone to. 

\paragraph{Transfer Learning:}
Because we are using an actor-critic learning method, we also studied the possibility of using the value functions for \TL. 
We did not discover any empirical evidence that this assisted the learning process. 
When transferring to a new task, the state distribution has changed and the reward function may be completely different. 
This makes it unlikely that the value function will be accurate on this new task. 
In addition, value functions are in general easier and faster to learn than policies, implying that value function reuse is less important to transfer.
We also find that helpfulness of \TL depends on not only the task difficulty but the reward function as well. 
Two tasks may overlap in state space but the area they overlap could be easily reachable. 
In this case \TL may not give significant benefit because the overall \RL problem is easy. 
The greatest benefit is gained from \TL when the state space that overlaps for two tasks is difficult to reach 
and in that difficult to reach area is where the highest rewards are achieved.

\subsection{Limitations:}
Once integrated, the skills for our locomotion tasks are self-selecting based on their context, i.e., the knowledge of the upcoming terrain.
It may be that other augmentation and distillation strategies are better for situations where either the reward functions are different or a one-hot vector is used to select the currently active expert.
In our transfer learning results we could be over fitting the initial \expert for the particular task it was learning. 
Making it more challenging for the policy to learn a new task, resulting in \emph{negative transfer}. 
After learning many new tasks the previous tasks may not receive a large enough potion of the \distillationText training process to preserve the \experts skill well enough. How best to chose which data should be trained on next to best preserve the 
behaviour of \experts is a general problem with multi-task learning.
\textit{Distillation} treats all tasks equally independent of their reward. 
This can result in very low value tasks, receiving potentially more distribution than desired and high value tasks receiving not enough.
We have not needed the use a \textit{one-hot} vector to indicate what task the \agent is performing.
We want the \agent to be able to recognize which task it is given but we do realize that some tasks could be too similar to differentiate, such as, walking vs jogging on flat ground.

\subsection{Future Work:}

It would be interesting to develop a method to prioritize tasks during the distillation step. 
This could assist the agent with forgetting issues or help with relearning tasks.
While we currently use the \MSE to pull the distributions of \student policies in line with \expert polices for \distillationText,
better distance metrics would likely be helpful. 
Previous methods have used KL Divergence in the discrete action space domain where the state-action value function encodes the policy, e.g., as with \DQN. 
In this work we do not focus on producing the best policy from a \emph{mixture of experts}, 
but instead we match the distributions from a number of \experts. 
The difference is subtle but in practice it can be more challengine to balance many \experts with respect to their reward functions.
It could also be beneficial to use a KL penalty while performing \distillationText, i.e., 
something similar to the work in~\citep{teh2017distral} in order to keep the policy from changing too rapidly during training.

\section{Conclusion}

We have proposed and evaluated a method for the progressive learning and integration (via distillation) of
motion skills. The method exploits transfer learning to speed learning of new skills, along with input injection where needed,
as well as continuous-action \distillationText, using DAGGER-style learning.
This compares favorably to baselines consisting of learning all skills together, or learning all
the skills individually before integration. We believe that there remains much to learned about the best training and integration
methods for movement skill repertoires, as is also reflected in the human motor learning literature.

%% file: tex/appendix.tex
\section{Appendix}
\label{sec:appendix}

\subsection{Network Models}
We used two different Network models for the experiments in this paper.
The first model is a \textit{blind} model that does not have any terrain features.
The \textit{blind} policy is a Neural Network with $2$ hidden layers ($512 \times 256$) with ReLU activations.
The output layer of the policy network has linear activations.
The network used for the value function has the same design except there is $1$ output on the final layer.
This design is used for the \Flat and \incline tasks.

We augment the \textit{blind} network design by adding features for terrain to create an \agent with \textit{sight}.
This network with \textit{terrain features} has a single convolution layer with $8$ filters of width $3$.
This constitutional layer is followed by a dense layer of $32$ units.
The dense layer is then concatenated twice, once along each of the original two hidden layers in the \textit{blind} version of the policy.

\subsection{Hyper Parameters and Training}

The policy network models a Gaussian distribution by outputting a state dependant mean. We use a state independent standard deviation that normalized with respect to the action space and multiplied by $0.1$.
We also use a version of epsilon greedy exploration where with $\epsilon$ probability an exploration action is generated.
For all of our experiments we linearly anneal $\epsilon$ from $0.2$ to $0.1$ in $100,000$ iterations and leave it from that point on.
Each training simulation takes approximately $5$ hours across $8$ threads.
For network training we use \SGD with momentum.
During the distillation step we use gradually anneal the probability of selecting an \expert action from $1$ to $0$ over $10,000$ iterations.

For the evaluation of each model on a particular task we use the average reward achieved by the \agent over at most $100$ seconds of simulation time.
We average this over running the \agent over a number of randomly generated simulation runs.

\subsubsection{\changes{Distillation}}
\label{subsection:distillation-details}
\changes{
For each of the distillation steps we initialize the policy from the most recently trained policy.
This policy has seen all of the tasks thus far but may have overfit the most recent tasks.
We us a version of the DAGGER algorithm for the distillation process~\citep{DBLP:journals/corr/abs-1011-0686}.
We anneal from selecting actions from the \expert polices to selecting actions from the \student policy
The probability of selecting an action from the \expert is annealed to near zero after $10,000$ training updates.
We still add exploration noise to the policies when generating actions to take in the simulation. This is also annealed along with the probability of selecting from the \expert policy.
The actions used for training always come from the \expert policy. Although some actions are applied in the simulation from the \student, during a training update those actions will be replaced with ones from the proper \expert.
The \expert used to generate actions for tasks $0-i$ is $\policySymbol_{i}$ and the \expert used to generate action for task $i+1$ is $\policySymbol_{i+1}$. We keep around at most $2$ policies at any time.
}



\subsection{Input Features and Injection}
\label{subsection:input-feature-injection}

In order to add additional input features to the policy network we construct a new network.
This new network has a portion of it that is the same design as the previous network plus additional parameters.
First we initialize the new network with random parameters.
Then we copy over the values from the previous network into the new one for the portion of the network design that matches the old.
Then the weight for the layers that connect the old portion of the network to the new are set to $0$.
This will allow the network to preserve the previous distribution it modeled.
Having the parameters from the old network will also help generate gradients to train the new $0$ valued network parameters.
\changes{
We use feature injection to assist the learning method with differentiating between different states. 
For example, it could be challenging to discover the difference between the \Flat and \incline tasks using only the character features.
Therefore, we add new terrain features to allow the controller to better differentiate between these two different tasks.
}

\begin{figure}[ht!]
\centering
\subcaptionbox{\label{fig:input-features}}{ \includegraphics[trim={0.0cm 2.5cm 0.0cm 3.0cm},clip,width=0.48\columnwidth]{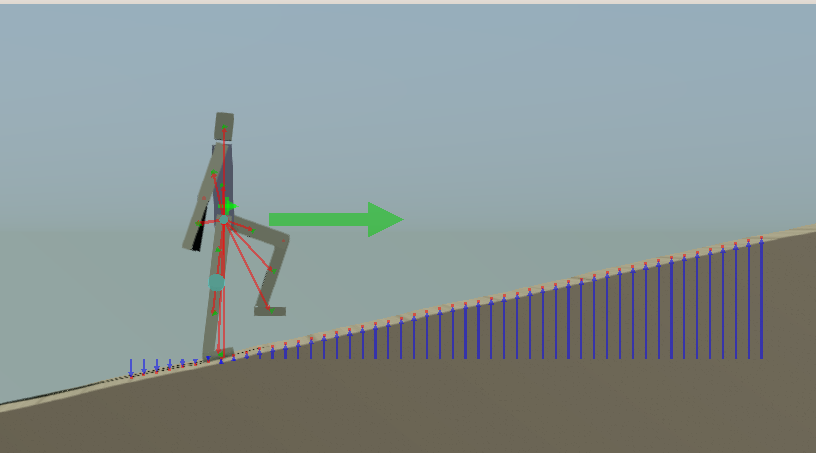}} 
\subcaptionbox{\label{fig:inject-input-terrain}}{ \includegraphics[trim={0.0cm 0.0cm 0.0cm 0.0cm},clip,width=0.48\columnwidth]{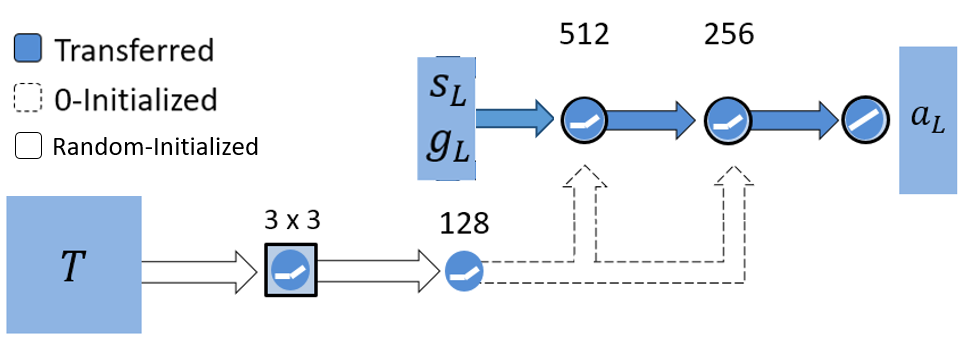}}
\caption{
\changes{(a) The input features include both the character state shown as the red lines between the root of the character and the character's links and the terrain features shown as the blue arrows along the ground.
(b) A diagram of method used to inject additional state features for the terrain.
}
}
\label{fig:inject-input}
\end{figure}


\subsection{\TL-Only Baseline}
\label{subsection:tl-only-baseline}
\begin{figure}[!ht]
\centering 
\subcaptionbox{\label{fig:TL-only-steps} \steps}{ \includegraphics[width=0.32\columnwidth]{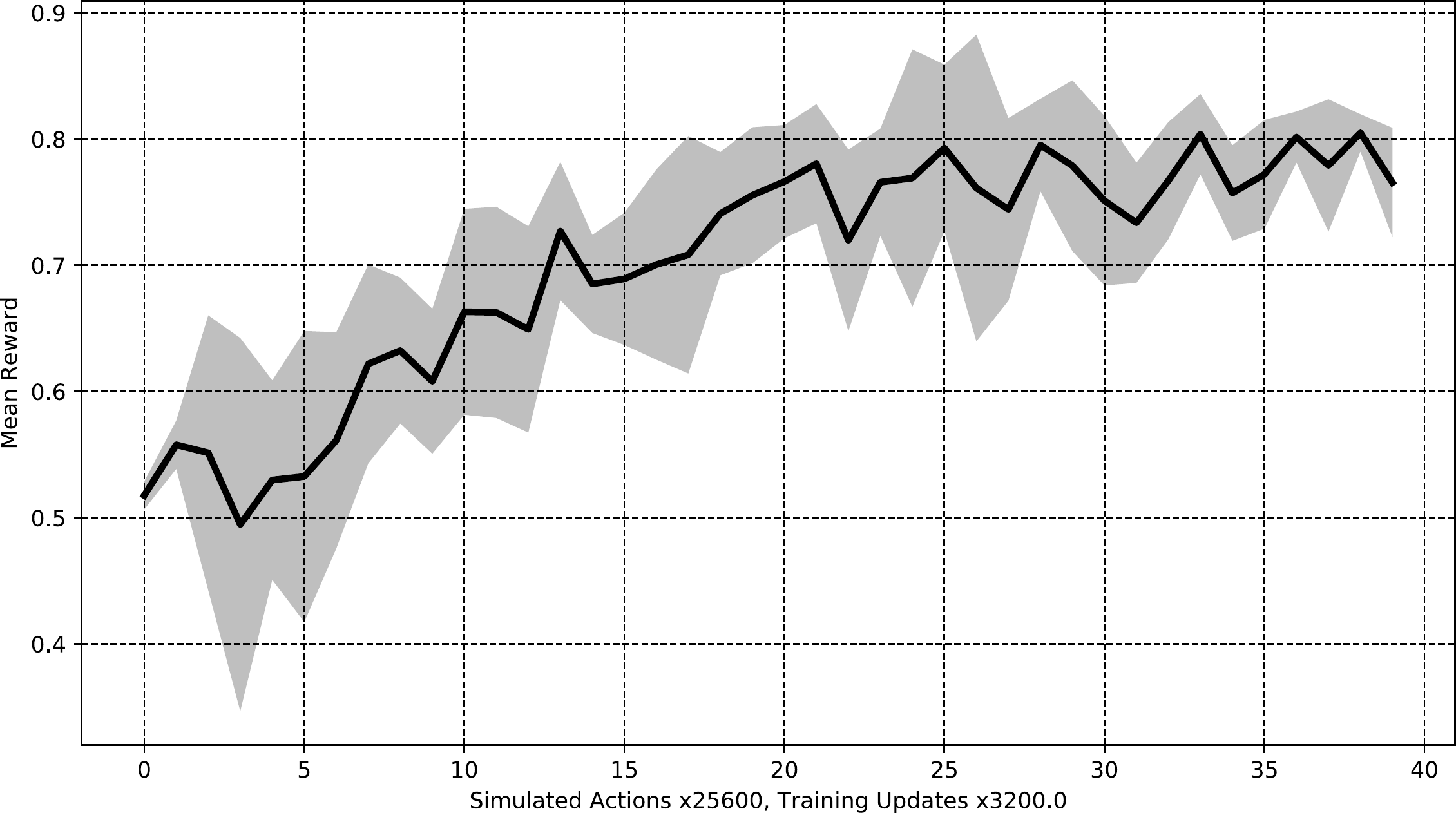}}
\subcaptionbox{\label{fig:TL-only-slopes} \slopes}{ \includegraphics[width=0.32\columnwidth]{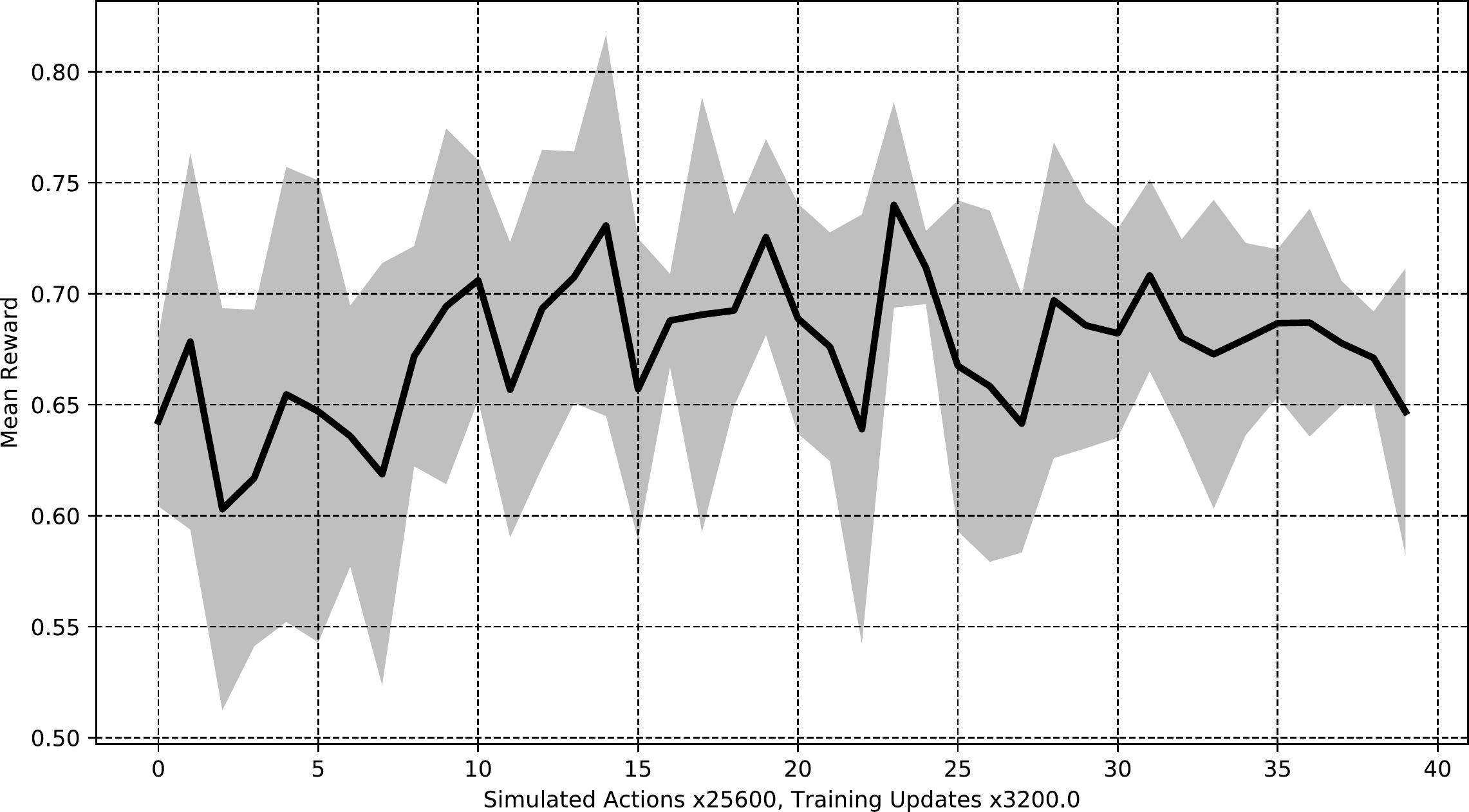}}
\subcaptionbox{\label{fig:TL-only-gaps} \gaps}{ \includegraphics[width=0.32\columnwidth]{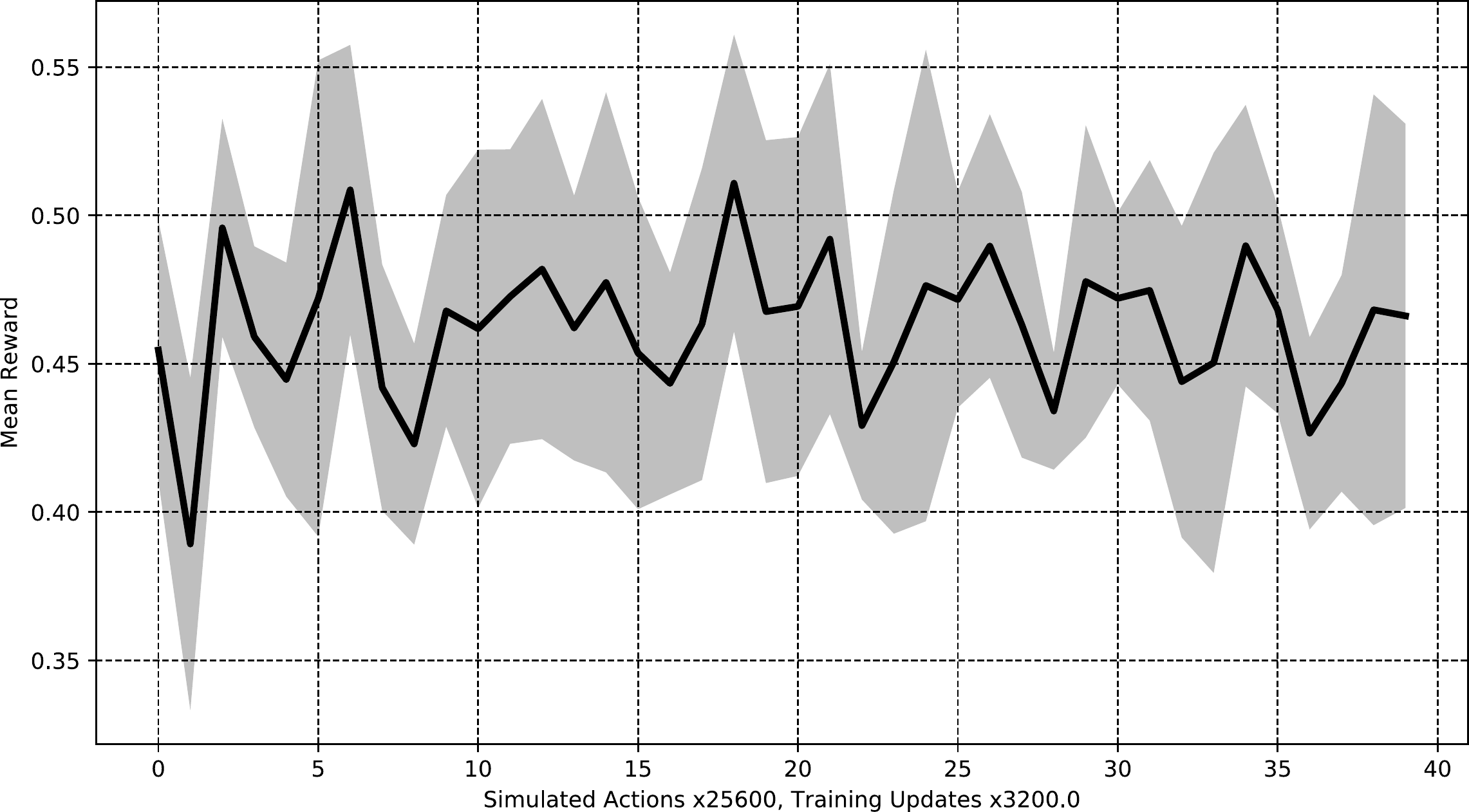}} 
\caption{
\changes{Transfer learning only baselines for each of the new tasks.}
}
\label{fig:TL-only}
\end{figure}

\changes{We also evaluate a baseline where we \TL for all tasks. 
In this baseline \TL is performed for a number of tasks and then \distillationText is used to combined these many learned skills.
This method can be considered a version of \PLAID where tasks are learned in groups and after some number of tasks, a collection of policies/skills are distilled together.
In~\refFigure{fig:TL-only} the learning curves for the \TLOnly baseline are given.
The \TLOnly method learns new tasks well.
We do not show the \incline tasks as the two methods are the same up to starting the \steps tasks.
In~\refTable{fig:TL-only-comparison} the amount of forgetting is compared between methods.
To compare the amount of forgetting between \TLOnly and \PLAID we show the relative loss in average reward between the original policy trained for the tasks \steps and \slopes and the final polices for each method on \gaps.
The \TLOnly method shows a larger drop in policy performance corresponding to a large amount of forgeting compared to \PLAID, in particular for the more complex tasks \steps and \slopes.
Interestingly, the final distllation step for \TLOnly appears to reduce the performance of the policy. We believe this is related to the final distillation step being more challenging than performing a simpler distillation after each new task.
Note that we only compare these two steps because the process for the first two tasks for \PLAID and \TLOnly are the same.
A comparison of the average rewards for the final policies are given in~\refTable{figure:final-evaluation}.
}

\begin{table}
	\centering
	\begin{tabular}{|p{3.5cm}||p{1.0cm}|p{1.25cm}|p{1.25cm}|p{1.25cm}|p{1.25cm}|p{1.25cm}|}
		\hline
		Tasks & \Flat & \incline & \steps & \slopes & \gaps & \textbf{average} \\ \hline \hline
		\PLAID & \num{0.89072313686} & \num{0.7997847458} & \num{0.66610235084} & \textbf{\num{0.60244157756}} &  \num{0.5289199521} & \num{0.6975943526} \\ \hline
		\TLOnly & \num{0.79047187764} & \num{0.66154053498} & \num{0.61528060036} & \num{0.54316894964} & \num{0.62609913758} & \num{0.64731222} \\ \hline
		\TLOnly (with Distill) & \num{0.90305467034} & \num{0.71922665866} & \num{0.7807573524} & \num{0.6714701784} & \num{0.54259102838} & \num{0.7234199776} \\ \hline
		\MultiTasker & \num{0.84424672714} & \num{0.75652856638} & \num{0.67734455282} & \num{0.65631263654} & \num{0.50404415568} & \num{0.6876953277} \\ \hline
	\end{tabular}
	\caption{\label{figure:final-evaluation}
			Final average reward for each method.
			Higher is better.
			Here, the final policy is after training on \gaps.
			the \PLAID method achieves on average higher values across tasks.
	}
\end{table}
\subsection{Agent design}

The \agent used in the simulation models the dimensions and masses of the average adult.
The size of the character state is $50$ parameters that include the relative position and velocity of the links in the \agent (\refFigure{fig:input-features}).
The action space consists of $11$ parameters that indicate target joint positions for the \agent.
The target joint positions (pd-targets) are turned into joint torques via proportional derivative controllers at each joint.

The reward function for the agent consists of 3 primary terms. 
The first is a velocity term the rewards the \agent for going at velocity of \valueWithUnits{$1$}{m/s}
The second term is the difference between the pose of the \agent and the current pose of a kinematic character controlled via a motion capture clip.
The difference between the \agent and the clip consists of the rotational difference between each corresponding joint and the difference in angular velocity.
The angular velocity for the clip is approximated via finite differences between the current pose of the clip and it's last pose.
The last term is an L2 penalty on the torques generated by the \agent to help reduce spastic motions.
We also impose torque limits on the joints to reduce unrealistic behaviour, limits: Hips $150$, knees $125$, ankles $100$, shoulders $100$, elbows $75$ and neck \valueWithUnits{$50$}{N/m}.

\label{sec:training-details-on-training}

\paragraph{Terrain Types}
All terrain types are randomly generated per episode, except for the \Flat terrain.
The \incline terrain is slanted and the slant of the terrain is randomly sampled between $20$ and $25$ degrees.
The \steps terrain consists of flat segments with widths randomly sampled from \valueWithUnits{$1.0$}{m} to \valueWithUnits{$1.5$}{m} followed by sharp steps that have randomly generated heights between \valueWithUnits{$5$}{cm} and \valueWithUnits{$15$}{cm}.
The \slopes terrain is randomly generated by updating the slope of the previous point in the ground with a value sampled from $-20$ and $20$ degrees to generate a new portion of the ground every \valueWithUnits{$10$}{cm}.
The \gaps terrain generate gaps of width \valueWithUnits{$25$ - $30$}{cm} separated by flat segments of widths sampled from \valueWithUnits{$2.0$}{m} to \valueWithUnits{$2.5$}{m}.
The \mixed terrain is a combination of the above terrains where a portion is randomly chosen from the above terrain types.


\begin{figure}[!ht]
\centering 
\subcaptionbox{\label{fig:distilation-environemnts-flat} \Flat}{ \includegraphics[trim={7.5cm 3.5cm 7.5cm 3.5cm},clip,width=0.32\columnwidth]{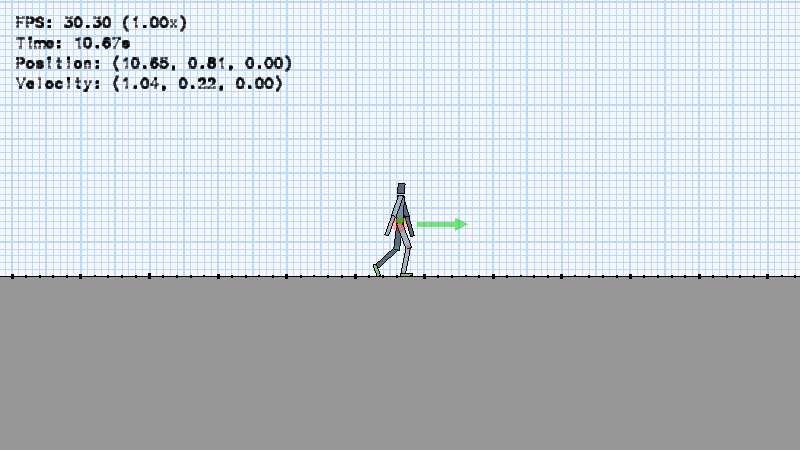}}
\subcaptionbox{\label{fig:distilation-environemnts-incline} \incline}{ \includegraphics[trim={7.5cm 3.5cm 7.5cm 3.5cm},clip,width=0.32\columnwidth]{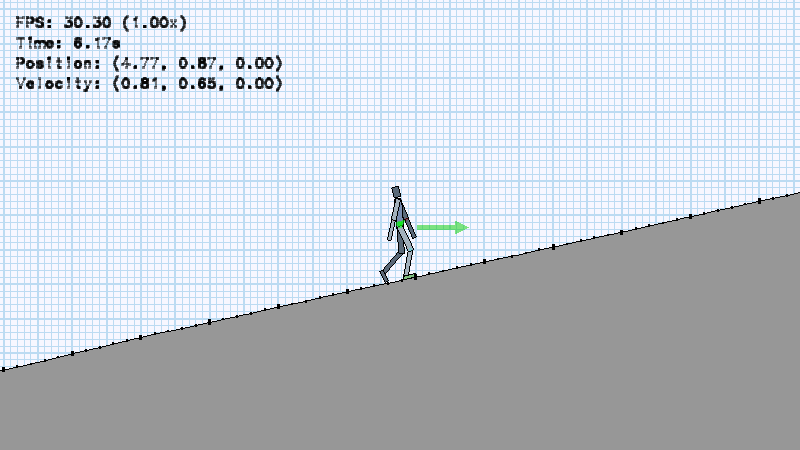}}
\subcaptionbox{\label{fig:distilation-environemnts-steps} \steps}{ \includegraphics[trim={7.5cm 3.5cm 7.5cm 3.5cm},clip,width=0.32\columnwidth]{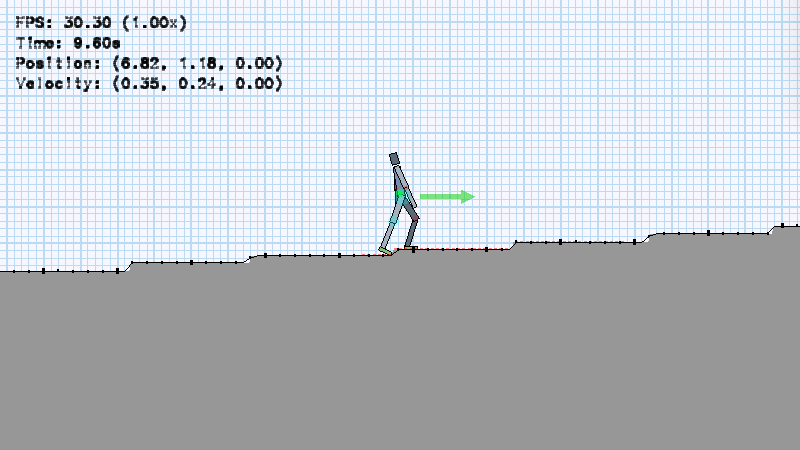}} \\
\subcaptionbox{\label{fig:distilation-environemnts-slopes} \slopes}{ \includegraphics[trim={2.5cm 0.0cm 2.5cm 3.5cm},clip,width=0.32\columnwidth]{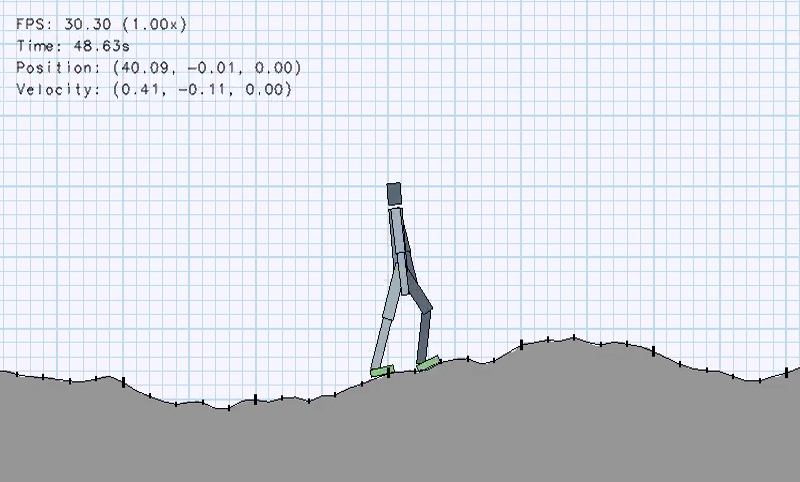}}
\subcaptionbox{\label{fig:distilation-environemnts-gaps} \gaps}{ \includegraphics[trim={2.5cm 0.0cm 2.5cm 3.5cm},clip,width=0.32\columnwidth]{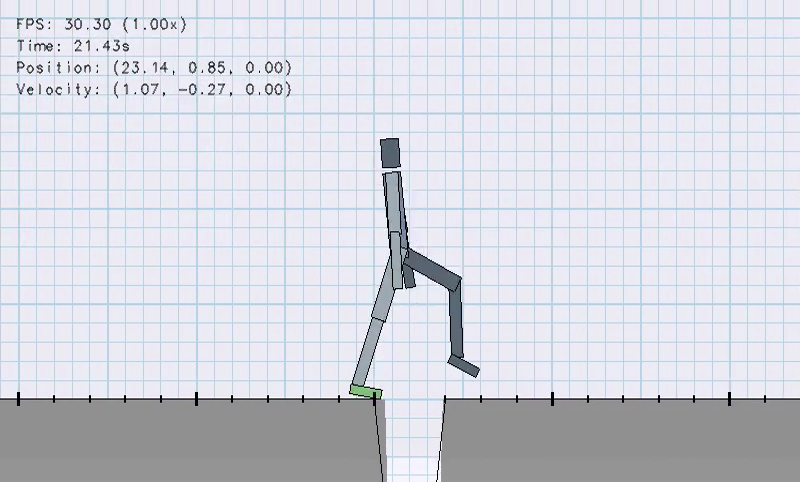}} 
\subcaptionbox{\label{fig:distilation-environemnts-mixed} \mixed}{ \includegraphics[trim={2.5cm 0.0cm 2.5cm 3.5cm},clip,width=0.32\columnwidth]{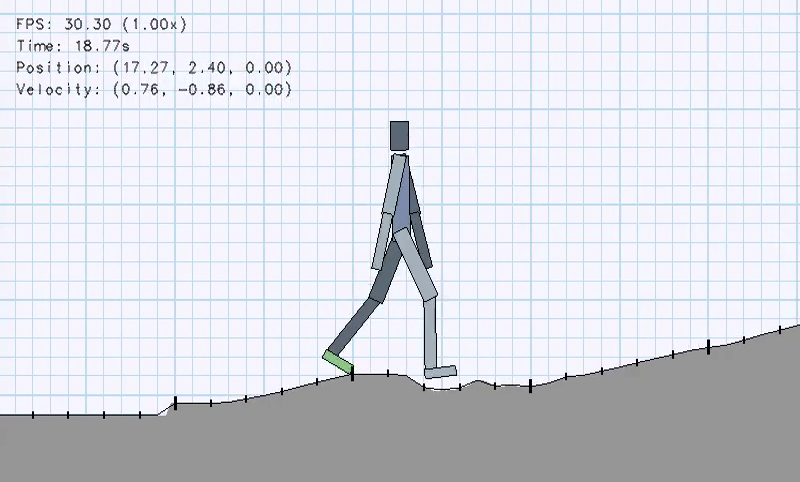}} \\
\caption{
The environments used to evaluate \progRL.
}
\label{fig:distilation-environemnts}
\end{figure}

\begin{figure}[!ht]
\centering 
\subcaptionbox{\label{fig:mixed-environemnts-1} }{\includegraphics[trim={10.5cm 2.5cm 10.5cm 2.5cm},clip,width=0.13\columnwidth]{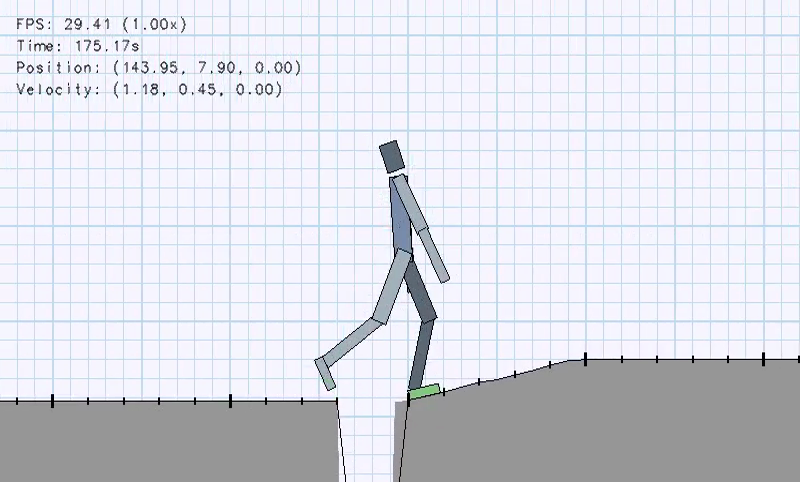}}
\subcaptionbox{\label{fig:mixed-environemnts-2} }{ \includegraphics[trim={10.5cm 2.0cm 10.5cm 3.0cm},clip,width=0.13\columnwidth]{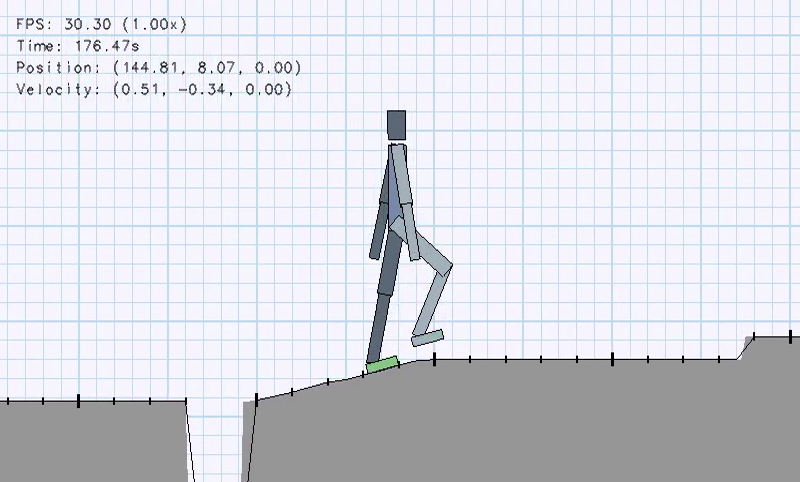}} 
\subcaptionbox{\label{fig:mixed-environemnts-3} }{ \includegraphics[trim={10.5cm 2.0cm 10.5cm 3.0cm},clip,width=0.13\columnwidth]{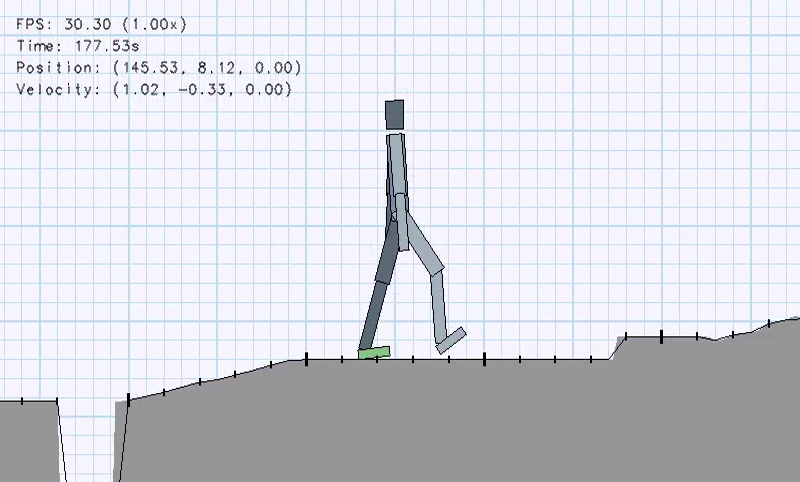}}
\subcaptionbox{\label{fig:mixed-environemnts-4} }{ \includegraphics[trim={10.5cm 2.0cm 10.5cm 3.0cm},clip,width=0.13\columnwidth]{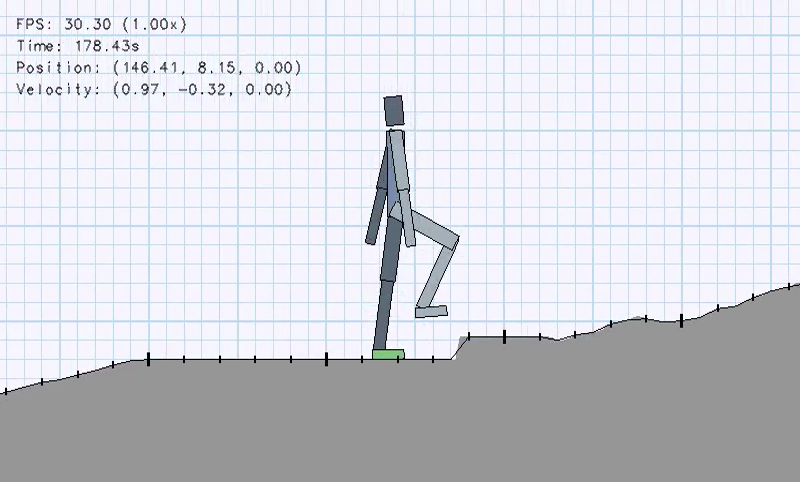}} 
\subcaptionbox{\label{fig:mixed-environemnts-5} }{ \includegraphics[trim={10.5cm 1.0cm 10.5cm 4.0cm},clip,width=0.13\columnwidth]{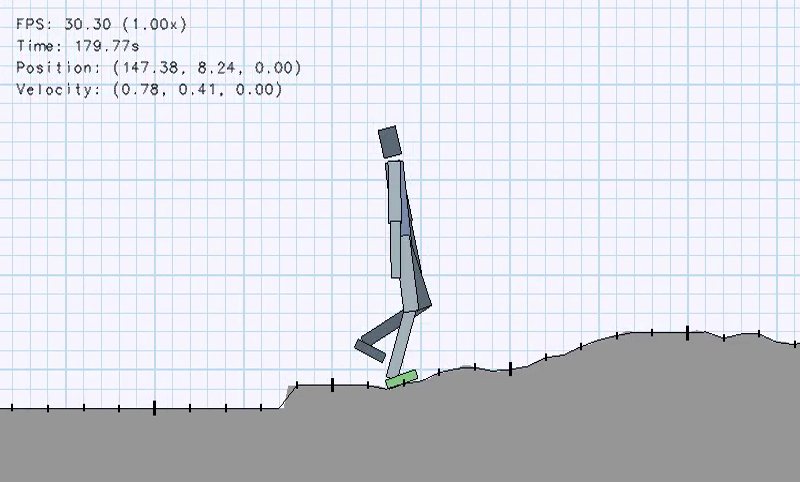}} 
\subcaptionbox{\label{fig:mixed-environemnts-6} }{ \includegraphics[trim={10.5cm 2.0cm 10.5cm 3.0cm},clip,width=0.13\columnwidth]{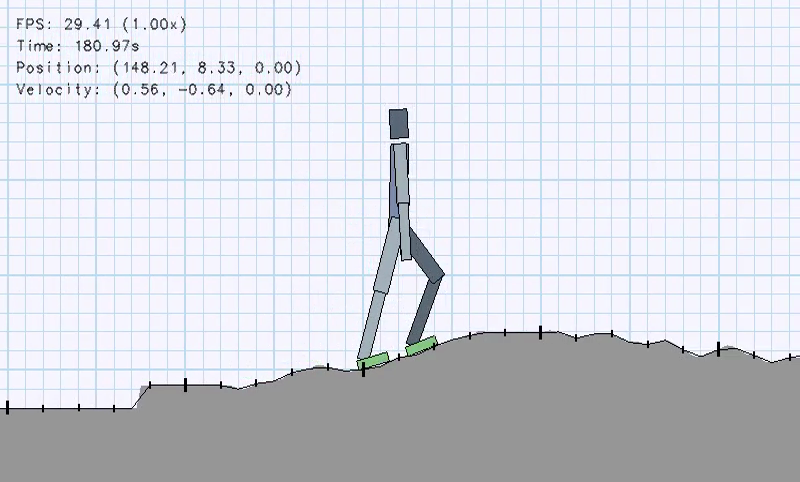}} 
\subcaptionbox{\label{fig:mixed-environemnts-7} }{ \includegraphics[trim={10.5cm 1.0cm 10.5cm 4.0cm},clip,width=0.13\columnwidth]{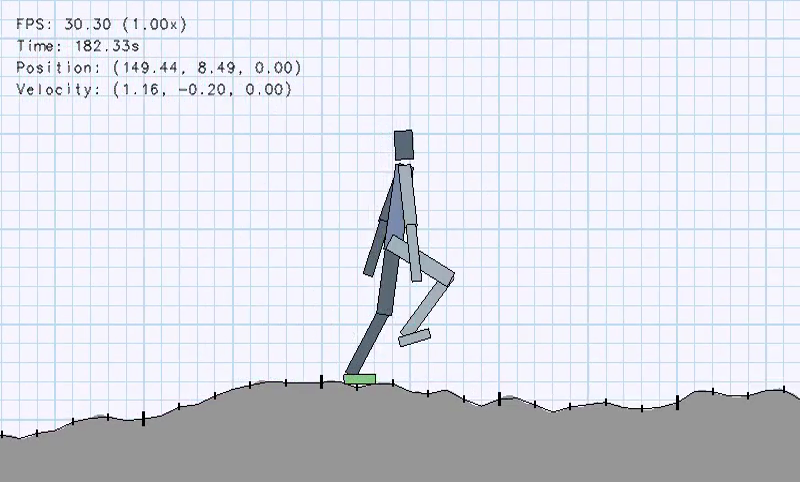}} 
\caption{
Still frame shots of the \pdcontrollerText traversing the \mixed environment.
}
\label{fig:mixed-environemnts}
\end{figure}

\subsection{Multitasker}

\begin{figure}[ht!]
\centering
\subcaptionbox{\label{fig:multitasker-evalution-multiple-policies}}{ \includegraphics[trim={0.0cm 0.0cm 0.0cm 1.35cm},clip,width=0.42\columnwidth]{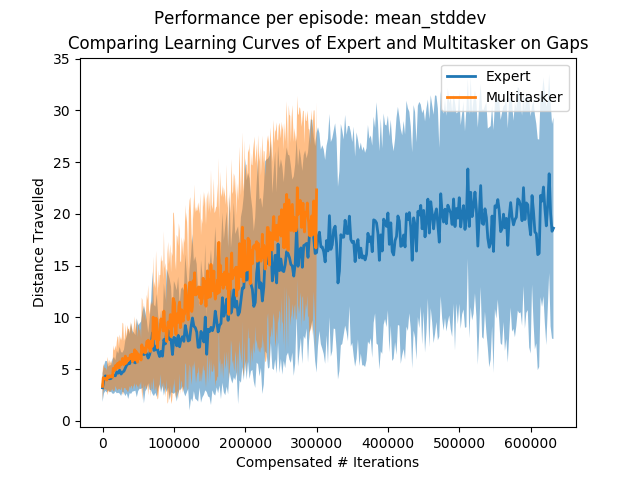}}
\caption{
(a) Shows that the \MultiTasker can learn faster on \steps, \Flat and \incline than \progRL (expert) learning the single task \steps with \TL.
}
\label{fig:multitasker-evalution}
\end{figure}

In certain cases the \MultiTasker can learn new task faster than \progRL. 
In~\refFigure{fig:multitasker-evalution-multiple-policies} we present the \MultiTasker and compare it to \progRL.
In this case the \MultiTasker splits its training time across multiple tasks, here we compare the two methods with respect to the time spent learning on the single \textit{new} task.
This is a good baseline to compare our method against but in some ways this is not fair.
If the real measure of how efficient a learning method is the number of simulation samples that are needed to learn would fall far behind as the \MultiTasker needs to train across all tasks to gain the benefits of improving a single task without forgetting the old tasks.

%% file: paper.bbl
\begin{thebibliography}{32}
\providecommand{\natexlab}[1]{#1}
\providecommand{\url}[1]{\texttt{#1}}
\expandafter\ifx\csname urlstyle\endcsname\relax
  \providecommand{\doi}[1]{doi: #1}\else
  \providecommand{\doi}{doi: \begingroup \urlstyle{rm}\Url}\fi

\bibitem[Bellemare et~al.(2013)Bellemare, Naddaf, Veness, and
  Bowling]{bellemare2013arcade}
Marc~G Bellemare, Yavar Naddaf, Joel Veness, and Michael Bowling.
\newblock The arcade learning environment: An evaluation platform for general
  agents.
\newblock \emph{J. Artif. Intell. Res.(JAIR)}, 47:\penalty0 253--279, 2013.

\bibitem[Bengio et~al.(2015)Bengio, Vinyals, Jaitly, and
  Shazeer]{NIPS2015_5956}
Samy Bengio, Oriol Vinyals, Navdeep Jaitly, and Noam Shazeer.
\newblock Scheduled sampling for sequence prediction with recurrent neural
  networks.
\newblock In C.~Cortes, N.~D. Lawrence, D.~D. Lee, M.~Sugiyama, and R.~Garnett
  (eds.), \emph{Advances in Neural Information Processing Systems 28}, pp.\
  1171--1179. Curran Associates, Inc., 2015.
\newblock URL
  \url{http://papers.nips.cc/paper/5956-scheduled-sampling-for-sequence-prediction-with-recurrent-neural-networks.pdf}.

\bibitem[Chen et~al.(2015)Chen, Goodfellow, and Shlens]{chen2015net2net}
Tianqi Chen, Ian Goodfellow, and Jonathon Shlens.
\newblock Net2net: Accelerating learning via knowledge transfer.
\newblock \emph{arXiv preprint arXiv:1511.05641}, 2015.

\bibitem[Devin et~al.(2017)Devin, Gupta, Darrell, Abbeel, and
  Levine]{devin2017learning}
Coline Devin, Abhishek Gupta, Trevor Darrell, Pieter Abbeel, and Sergey Levine.
\newblock Learning modular neural network policies for multi-task and
  multi-robot transfer.
\newblock In \emph{Robotics and Automation (ICRA), 2017 IEEE International
  Conference on}, pp.\  2169--2176. IEEE, 2017.

\bibitem[Finn et~al.(2017)Finn, Abbeel, and Levine]{finn2017model}
Chelsea Finn, Pieter Abbeel, and Sergey Levine.
\newblock Model-agnostic meta-learning for fast adaptation of deep networks.
\newblock \emph{arXiv preprint arXiv:1703.03400}, 2017.

\bibitem[Heess et~al.(2016)Heess, Wayne, Tassa, Lillicrap, Riedmiller, and
  Silver]{DBLP:journals/corr/HeessWTLRS16}
Nicolas Heess, Gregory Wayne, Yuval Tassa, Timothy~P. Lillicrap, Martin~A.
  Riedmiller, and David Silver.
\newblock Learning and transfer of modulated locomotor controllers.
\newblock \emph{CoRR}, abs/1610.05182, 2016.
\newblock URL \url{http://arxiv.org/abs/1610.05182}.

\bibitem[Kirkpatrick et~al.(2017)Kirkpatrick, Pascanu, Rabinowitz, Veness,
  Desjardins, Rusu, Milan, Quan, Ramalho, Grabska-Barwinska, Hassabis, Clopath,
  Kumaran, and Hadsell]{Kirkpatrick28032017}
James Kirkpatrick, Razvan Pascanu, Neil Rabinowitz, Joel Veness, Guillaume
  Desjardins, Andrei~A. Rusu, Kieran Milan, John Quan, Tiago Ramalho, Agnieszka
  Grabska-Barwinska, Demis Hassabis, Claudia Clopath, Dharshan Kumaran, and
  Raia Hadsell.
\newblock Overcoming catastrophic forgetting in neural networks.
\newblock \emph{Proceedings of the National Academy of Sciences}, 114\penalty0
  (13):\penalty0 3521--3526, 2017.
\newblock \doi{10.1073/pnas.1611835114}.
\newblock URL \url{http://www.pnas.org/content/114/13/3521.abstract}.

\bibitem[Kulkarni et~al.(2016)Kulkarni, Narasimhan, Saeedi, and
  Tenenbaum]{HDeepRL}
Tejas~D Kulkarni, Karthik Narasimhan, Ardavan Saeedi, and Josh Tenenbaum.
\newblock Hierarchical deep reinforcement learning: Integrating temporal
  abstraction and intrinsic motivation.
\newblock In \emph{Advances in Neural Information Processing Systems 29}, pp.\
  3675--3683. 2016.

\bibitem[Lamb et~al.(2016)Lamb, ALIAS PARTH~GOYAL, Zhang, Zhang, Courville, and
  Bengio]{NIPS2016_6099}
Alex~M Lamb, Anirudh~Goyal ALIAS PARTH~GOYAL, Ying Zhang, Saizheng Zhang,
  Aaron~C Courville, and Yoshua Bengio.
\newblock Professor forcing: A new algorithm for training recurrent networks.
\newblock In D.~D. Lee, M.~Sugiyama, U.~V. Luxburg, I.~Guyon, and R.~Garnett
  (eds.), \emph{Advances in Neural Information Processing Systems 29}, pp.\
  4601--4609. Curran Associates, Inc., 2016.
\newblock URL
  \url{http://papers.nips.cc/paper/6099-professor-forcing-a-new-algorithm-for-training-recurrent-networks.pdf}.

\bibitem[Li \& Hoiem(2016)Li and Hoiem]{DBLP:journals/corr/LiH16e}
Zhizhong Li and Derek Hoiem.
\newblock Learning without forgetting.
\newblock \emph{CoRR}, abs/1606.09282, 2016.
\newblock URL \url{http://arxiv.org/abs/1606.09282}.

\bibitem[Martinez et~al.(2017)Martinez, Black, and
  Romero]{DBLP:journals/corr/MartinezBR17}
Julieta Martinez, Michael~J. Black, and Javier Romero.
\newblock On human motion prediction using recurrent neural networks.
\newblock \emph{CoRR}, abs/1705.02445, 2017.
\newblock URL \url{http://arxiv.org/abs/1705.02445}.

\bibitem[Merel et~al.(2017)Merel, Tassa, Srinivasan, Lemmon, Wang, Wayne, and
  Heess]{merel2017learning}
Josh Merel, Yuval Tassa, Sriram Srinivasan, Jay Lemmon, Ziyu Wang, Greg Wayne,
  and Nicolas Heess.
\newblock Learning human behaviors from motion capture by adversarial
  imitation.
\newblock \emph{arXiv preprint arXiv:1707.02201}, 2017.

\bibitem[Pan \& Yang(2010)Pan and Yang]{5288526}
S.~J. Pan and Q.~Yang.
\newblock A survey on transfer learning.
\newblock \emph{IEEE Transactions on Knowledge and Data Engineering},
  22\penalty0 (10):\penalty0 1345--1359, Oct 2010.
\newblock ISSN 1041-4347.
\newblock \doi{10.1109/TKDE.2009.191}.

\bibitem[Parisotto et~al.(2015)Parisotto, Ba, and
  Salakhutdinov]{parisotto2015actor}
Emilio Parisotto, Jimmy~Lei Ba, and Ruslan Salakhutdinov.
\newblock Actor-mimic: Deep multitask and transfer reinforcement learning.
\newblock \emph{arXiv preprint arXiv:1511.06342}, 2015.

\bibitem[Peng \& van~de Panne(2016)Peng and van~de Panne]{PengP16}
Xue~Bin Peng and Michiel van~de Panne.
\newblock Learning locomotion skills using deeprl: Does the choice of action
  space matter?
\newblock \emph{CoRR}, abs/1611.01055, 2016.
\newblock URL \url{http://arxiv.org/abs/1611.01055}.

\bibitem[Peng et~al.(2017)Peng, Berseth, Yin, and Van
  De~Panne]{peng2017deeploco}
Xue~Bin Peng, Glen Berseth, Kangkang Yin, and Michiel Van De~Panne.
\newblock Deeploco: Dynamic locomotion skills using hierarchical deep
  reinforcement learning.
\newblock \emph{ACM Transactions on Graphics (TOG)}, 36\penalty0 (4):\penalty0
  41, 2017.

\bibitem[{Rajendran} et~al.(2015){Rajendran}, {Lakshminarayanan}, {Khapra},
  {Prasanna}, and {Ravindran}]{rajendran2017attend}
J.~{Rajendran}, A.~S. {Lakshminarayanan}, M.~M. {Khapra}, P~{Prasanna}, and
  B.~{Ravindran}.
\newblock Attend, adapt and transfer: Attentive deep architecture for adaptive
  transfer from multiple sources in the same domain.
\newblock \emph{arXiv preprint arXiv:1510.02879}, October 2015.

\bibitem[Roemmich \& Bastian(2015)Roemmich and Bastian]{roemmich2015two}
Ryan~T Roemmich and Amy~J Bastian.
\newblock Two ways to save a newly learned motor pattern.
\newblock \emph{Journal of neurophysiology}, 113\penalty0 (10):\penalty0
  3519--3530, 2015.

\bibitem[Ross et~al.(2010)Ross, Gordon, and
  Bagnell]{DBLP:journals/corr/abs-1011-0686}
St{\'{e}}phane Ross, Geoffrey~J. Gordon, and J.~Andrew Bagnell.
\newblock No-regret reductions for imitation learning and structured
  prediction.
\newblock \emph{CoRR}, abs/1011.0686, 2010.
\newblock URL \url{http://arxiv.org/abs/1011.0686}.

\bibitem[Rusu et~al.(2015)Rusu, Colmenarejo, Gulcehre, Desjardins, Kirkpatrick,
  Pascanu, Mnih, Kavukcuoglu, and Hadsell]{rusu2015policy}
Andrei~A Rusu, Sergio~Gomez Colmenarejo, Caglar Gulcehre, Guillaume Desjardins,
  James Kirkpatrick, Razvan Pascanu, Volodymyr Mnih, Koray Kavukcuoglu, and
  Raia Hadsell.
\newblock Policy distillation.
\newblock \emph{arXiv preprint arXiv:1511.06295}, 2015.

\bibitem[Rusu et~al.(2016)Rusu, Rabinowitz, Desjardins, Soyer, Kirkpatrick,
  Kavukcuoglu, Pascanu, and Hadsell]{Rusu2016}
Andrei~A. Rusu, Neil~C. Rabinowitz, Guillaume Desjardins, Hubert Soyer, James
  Kirkpatrick, Koray Kavukcuoglu, Razvan Pascanu, and Raia Hadsell.
\newblock {Progressive Neural Networks}.
\newblock \emph{arXiv}, 2016.
\newblock URL \url{http://arxiv.org/abs/1606.04671}.

\bibitem[Schmidhuber(2011)]{DBLP:journals/corr/abs-1112-5309}
J{\"{u}}rgen Schmidhuber.
\newblock {POWERPLAY:} training an increasingly general problem solver by
  continually searching for the simplest still unsolvable problem.
\newblock \emph{CoRR}, abs/1112.5309, 2011.
\newblock URL \url{http://arxiv.org/abs/1112.5309}.

\bibitem[Schulman et~al.(2016)Schulman, Moritz, Levine, Jordan, and
  Abbeel]{schulman2016high}
John Schulman, Philipp Moritz, Sergey Levine, Michael Jordan, and Pieter
  Abbeel.
\newblock High-dimensional continuous control using generalized advantage
  estimation.
\newblock In \emph{International Conference on Learning Representations (ICLR
  2016)}, 2016.

\bibitem[Shin et~al.(2017)Shin, Lee, Kim, and Kim]{shin2017continual}
Hanul Shin, Jung~Kwon Lee, Jaehong Kim, and Jiwon Kim.
\newblock Continual learning with deep generative replay.
\newblock \emph{arXiv preprint arXiv:1705.08690}, 2017.

\bibitem[Silver et~al.(2014)Silver, Lever, Heess, Degris, Wierstra, and
  Riedmiller]{silver2014deterministic}
David Silver, Guy Lever, Nicolas Heess, Thomas Degris, Daan Wierstra, and
  Martin Riedmiller.
\newblock Deterministic policy gradient algorithms.
\newblock In \emph{ICML}, 2014.

\bibitem[Srivastava et~al.(2012)Srivastava, Steunebrink, and
  Schmidhuber]{DBLP:journals/corr/abs-1210-8385}
Rupesh~Kumar Srivastava, Bas~R. Steunebrink, and J{\"{u}}rgen Schmidhuber.
\newblock First experiments with powerplay.
\newblock \emph{CoRR}, abs/1210.8385, 2012.
\newblock URL \url{http://arxiv.org/abs/1210.8385}.

\bibitem[Sutton et~al.(2000)Sutton, McAllester, Singh, and
  Mansour]{sutton2000policy}
Richard~S Sutton, David~A McAllester, Satinder~P Singh, and Yishay Mansour.
\newblock Policy gradient methods for reinforcement learning with function
  approximation.
\newblock In \emph{Advances in neural information processing systems}, pp.\
  1057--1063, 2000.

\bibitem[Taylor \& Stone(2009)Taylor and Stone]{taylor2009transfer}
Matthew~E Taylor and Peter Stone.
\newblock Transfer learning for reinforcement learning domains: A survey.
\newblock \emph{Journal of Machine Learning Research}, 10\penalty0
  (Jul):\penalty0 1633--1685, 2009.

\bibitem[Teh et~al.(2017)Teh, Bapst, Czarnecki, Quan, Kirkpatrick, Hadsell,
  Heess, and Pascanu]{teh2017distral}
Yee~Whye Teh, Victor Bapst, Wojciech~Marian Czarnecki, John Quan, James
  Kirkpatrick, Raia Hadsell, Nicolas Heess, and Razvan Pascanu.
\newblock Distral: Robust multitask reinforcement learning.
\newblock \emph{arXiv preprint arXiv:1707.04175}, 2017.

\bibitem[Tessler et~al.(2016)Tessler, Givony, Zahavy, Mankowitz, and
  Mannor]{Tessler2016}
Chen Tessler, Shahar Givony, Tom Zahavy, Daniel~J Mankowitz, and Shie Mannor.
\newblock {A Deep Hierarchical Approach to Lifelong Learning in Minecraft}.
\newblock \emph{arXiv}, pp.\  1--6, 2016.
\newblock URL \url{http://arxiv.org/abs/1604.07255}.

\bibitem[Van~Hasselt(2012)]{vanHasselt2012}
Hado Van~Hasselt.
\newblock Reinforcement learning in continuous state and action spaces.
\newblock In \emph{Reinforcement Learning}, pp.\  207--251. Springer, 2012.

\bibitem[Wolpert \& Flanagan(2016)Wolpert and
  Flanagan]{wolpert2016computations}
Daniel~M Wolpert and J~Randall Flanagan.
\newblock Computations underlying sensorimotor learning.
\newblock \emph{Current opinion in neurobiology}, 37:\penalty0 7--11, 2016.

\end{thebibliography}
